\documentclass[runningheads]{llncs}
\usepackage{graphicx}

\usepackage{tikz}
\usepackage{comment}
\usepackage{amsmath,amssymb} 
\usepackage{color}

\usepackage{graphicx}
\usepackage{amsmath}
\usepackage{amssymb}
\usepackage{booktabs}
\usepackage{enumitem}
\usepackage{verbatim}

\usepackage{subfigure}

\usepackage{wrapfig}
\usepackage{epsfig}
\usepackage[utf8]{inputenc} 
\usepackage[T1]{fontenc}    

\usepackage{siunitx}
\usepackage{cite}
\usepackage{ctable}
\usepackage{hhline}
\usepackage{booktabs}

\usepackage{csquotes}
\usepackage{pifont}
\usepackage{mathtools}
\usepackage{animate}
\newlength{\itemwidth}
\usepackage{anyfontsize}
\usepackage[super]{nth}

\usepackage{ctable}
\usepackage{soul}
\usepackage{multicol}
\usepackage{makecell}
\usepackage[normalem]{ulem}
\usepackage{multirow}
\usepackage{float}
\usepackage{stfloats}

\usepackage{stmaryrd}
\usepackage{mathstyle}

\usepackage{caption}
\usepackage{threeparttable} 
\usepackage[pagebackref,breaklinks,colorlinks]{hyperref}

\usepackage[capitalize]{cleveref}

\newcommand{\modelname}{\textsc{SmoothNet}}

\usepackage{color, colortbl}
\definecolor{Gray}{gray}{0.9}
\definecolor{Red}{RGB}{230, 57, 70}
\definecolor{Blue}{RGB}{0, 100, 148}

\crefname{section}{Sec.}{Secs.}
\Crefname{section}{Section}{Sections}
\Crefname{table}{Table}{Tables}
\crefname{table}{Tab.}{Tabs.}

\begin{document}
\pagestyle{headings}
\mainmatter
\def\ECCVSubNumber{1848}  

\title{SmoothNet: A Plug-and-Play Network for Refining Human Poses in Videos} 

\titlerunning{SmoothNet}
%
\author{Ailing Zeng$^{1}$ \and
Lei Yang$^{2}$ \and
Xuan Ju$^{1}$ \and
Jiefeng Li$^{3}$ \and
Jianyi Wang$^{4}$ \and
Qiang Xu$^{1}$}
\authorrunning{A. Zeng et al.}
%
\institute{$^{1}$The Chinese University of Hong Kong, $^{2}$
Sensetime Group Ltd., \\$^{3}$Shanghai Jiao Tong University, $^{4}$Nanyang Technological University \\
\email{\{alzeng, qxu\}@cse.cuhk.edu.hk}}

\maketitle

\begin{center}
    \centering
    \setlength{\tabcolsep}{0cm}
    \setlength{\itemwidth}{6cm}
    \begin{tabular}{cc}
    \animategraphics[width=\itemwidth, poster=14, loop, autoplay, final, nomouse, method=widget]{200}{img/teaser/aist_final/}{000040}{000079}
  
    &
  
    \animategraphics[width=\itemwidth, poster=63, loop, autoplay, final, nomouse, method=widget]{200}{img/teaser/downtown_final2/}{000001}{000080}
    \\
    (a) Rare Pose Refinement

    &
    (b) Occluded Pose Refinement

    \end{tabular}
    \captionof{figure}
    {State-of-the-art human pose/shape estimators (e.g., VIBE~\cite{kocabas2020vibe}) suffer from severe jitters on videos containing rarely seen or occluded poses, resulting in untrustworthy perceptions. We propose a novel plug-and-play temporal refinement network, \modelname, significantly alleviating this problem. \emph{Note that this is a video figure, best viewed with Acrobat Reader.}
  }
\label{fig:intro_comparision_vibe}
\end{center}
  

\begin{abstract}
When analyzing human motion videos, the output jitters from existing pose estimators are highly-unbalanced with varied estimation errors across frames. Most frames in a video are relatively easy to estimate and only suffer from slight jitters. In contrast, for rarely seen or occluded actions, the estimated positions of multiple joints largely deviate from the ground truth values for a consecutive sequence of frames, rendering significant jitters on them. 

To tackle this problem, we propose to attach a dedicated \emph{temporal-only} refinement network to existing pose estimators for jitter mitigation, named~\modelname. Unlike existing learning-based solutions that employ spatio-temporal models to co-optimize per-frame precision and temporal smoothness at all the joints, \modelname~models the natural smoothness characteristics in body movements by learning the long-range temporal relations of every joint without considering the noisy correlations among joints. With a simple yet effective motion-aware fully-connected network, \modelname~improves the temporal smoothness of existing pose estimators significantly and enhances the estimation accuracy of those challenging frames as a side-effect. Moreover, as a temporal-only model, a unique advantage of \modelname~is its strong transferability across various types of estimators, modalities, and datasets. Comprehensive experiments on \emph{five datasets} with \emph{eleven popular backbone networks} across \emph{2D and 3D pose estimation and body recovery tasks} demonstrate the efficacy of the proposed solution. Code is available at https://github.com/cure-lab/SmoothNet.

\end{abstract}

\section{Introduction}

Human pose estimation has broad applications such as motion analysis and human-computer interaction. While significant advancements have been achieved with novel deep learning techniques (e.g.,~\cite{newell2016stacked,chen2018cascaded,sun2019hrnet,zeng2020srnet,kanazawa2018end,kolotouros2019spin}), the estimation errors for rarely seen or occluded poses are still relatively high. 

When applying existing image-based pose estimators for video analysis, significant jitters occur on those challenging frames with large estimation errors as L1/L2 loss optimization is directionless. Moreover, they often last for a consecutive sequence of frames, causing untrustworthy perceptions (see \emph{Estimated Results} in Figure~\ref{fig:intro_comparision_vibe}). 
Various video-based pose estimators are proposed in the literature to mitigate this problem. Some use an end-to-end network that takes jitter errors into consideration~\cite{kanazawa2019learning,kocabas2020vibe,pavllo20193d,Wang2020MotionG3,choi2021tcmr}, while the rest smooth the estimation results with spatial-temporal refinement models~\cite{kim2021attention,luo2020meva,veges2020temporal} or low-pass filters~\cite{hyndman2011moving,van1992fft,kalman1960new,press1990savitzky,young1995gaus1d,casiez20121euro,Coskun2017LongSM,gauss2021spsmoothing}. These solutions, however, do not consider the highly-unbalanced nature of the jitters in the estimated poses, resulting in unsatisfactory performance. 

On the one hand, existing learning-based solutions  (including end-to-end and refinement networks) employ Spatio-temporal models to co-optimize per-frame precision and temporal smoothness at all the joints. This is a highly challenging task as jittering frames typically persist for a while, and they are associated with untrustworthy local temporal features and noisy correlation among the estimated joints. 
On the other hand, applying low-pass filters on each estimated joint with a long filtering window could reduce jitters to an arbitrarily small value. Nevertheless, such fixed temporal filters usually lead to considerable precision loss (e.g., over-smoothing) without prior knowledge about the distribution of human motions. 


Motivated by the above, this work proposes to attach a dedicated \emph{temporal-only} refinement network to existing 2D/3D pose estimators for jitter mitigation, named~\modelname. Without considering the noisy correlations (especially on jittering frames) among estimated joint positions, \modelname~models the natural smoothness characteristics in body movements in a data-driven manner. The main contributions of this paper include: 

\begin{itemize}

\item  We investigate the highly-unbalanced nature of the jitter problem with existing pose estimators and empirically show that significant jitters usually occur on a consecutive sequence of challenging frames with poor image quality, occlusion, or rarely seen poses.

\item To the best of our knowledge, this is the first data-driven temporal-only refinement solution for human motion jitter mitigation. Specifically, we design simple yet effective fully-connected networks with a long receptive field to learn the temporal relations of every joint for smoothing, and we show it outperforms other temporal models such as temporal convolutional networks (TCNs) and vanilla Transformers. 

\item As a temporal-only model, SmoothNet is a plug-and-play network with strong transferability across various types of estimators and datasets. 

\end{itemize}

\modelname~is conceptually simple yet empirically powerful. We conduct extensive experiments to validate its effectiveness and generalization capability on \emph{five datasets}, \emph{eleven backbone networks}, and \emph{three modalities} (2D/3D position and 6D rotation matrix~\cite{Zhou2019OnTC}). Our results show that \modelname~improves the temporal smoothness of existing pose estimators significantly and enhances the estimation accuracy of the challenging frames as a side-effect, especially for those video clips with severe pose estimation errors and long-term jitters.

\section{Preliminaries and Problem Definition}
\label{sec:method_definition}

\begin{figure}[t]	
	\subfigure[Sudden Jitters.] 
	{
		\begin{minipage}[t]{0.45\linewidth}
			\centering         
			\includegraphics[width=1.6in]{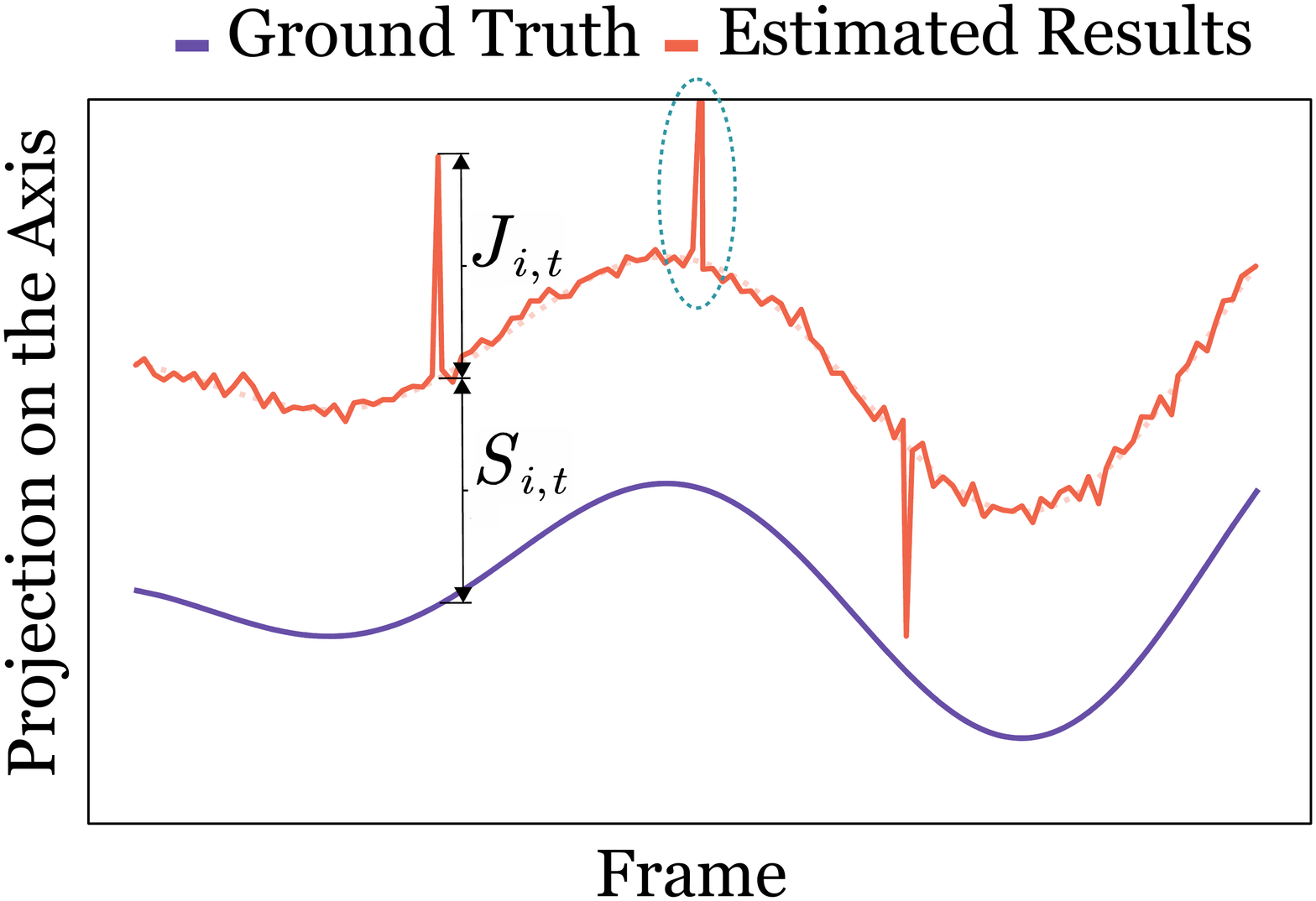}   
		\end{minipage}
	} 
    	\label{fig:jitter_point_large} 
    \subfigure[Long-term Jitters.] 
	{
		\begin{minipage}[t]{0.45\linewidth}
			\centering      
			\includegraphics[width=1.6in]{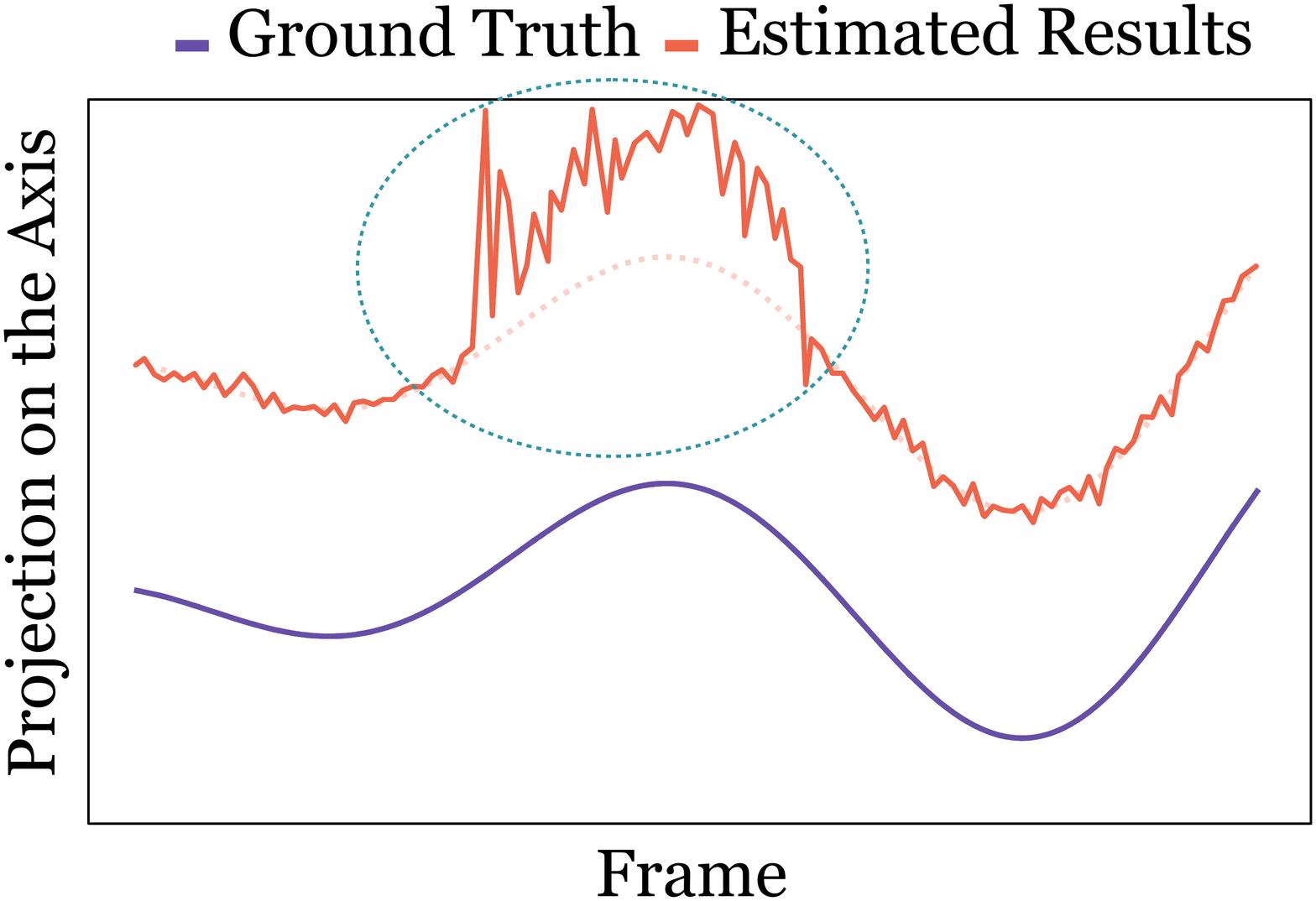}
		\end{minipage}
	}
    	\label{fig:jitter_seq_large}  

\caption{Two kinds of jitters are caused by pose estimation errors. The horizontal coordinate represents frames, and the vertical coordinate shows joint position values. Output errors are composed of jitter errors \emph{J} and biased errors \emph{S}.}

\label{fig:jitter} 
\end{figure}

\subsection{Human Pose Estimation}

For video-based human pose estimation, $L$ frames of a video $\mathbf{X}$ are inputs to the pose estimator $f$, and it outputs the estimated poses $\mathbf{\hat{Y}}\in \mathbb{R}^{L \times C}$, where $C = N \times D$.
$N$ is the number of keypoints associated with datasets, and $D$ denotes the dimensions of each keypoint (2D~\cite{newell2016stacked,chen2018cascaded,sun2019hrnet,li2021rle} or 3D~\cite{martinez2017simple,zhao2019semantic,pavllo20193d,zeng2020srnet,zeng2021learning}).
The above process can be simply formulated as $\mathbf{\hat{Y}} = f(\mathbf{X})$.
The estimator is trained in a supervised manner with the labeled ground truth $\mathbf{Y}\in \mathbb{R}^{L \times C}$.

\noindent \textbf{Key Evaluation Metrics.}
To evaluate the per-frame precision, the metric is the mean per joint position error (\emph{MPJPE}). To measure the smoothness or jitter errors, the metric is the mean per joint acceleration error (\emph{Accel}). 


\subsection{The Jitter Problem from Pose Estimators}
\label{sec:relat_video}

An ideal pose estimator that outputs accurate joint positions would not suffer from jitters. In other words, the jitter problem is caused by pose estimation errors, which can be divided into two parts: the \emph{jitter error} $\mathbf{J}$ between adjacent frames and the \emph{biased error} $\mathbf{S}$ between the ground truth and smoothed poses. In Figure~\ref{fig:jitter}, we differentiate sudden jitters and long-term jitters based on the duration of jitters. Moreover, according to the degree of jitters, existing jitters can be split into small jitters caused by inevitably inaccurate and inconsistent annotations in the training dataset (e.g.,~\cite{lin2014coco,andriluka2014mpii})
and large jitters caused by poor image quality, rare poses, or heavy occlusion. 

\begin{figure*}[t]
\begin{center}
\includegraphics[width=0.97\textwidth]{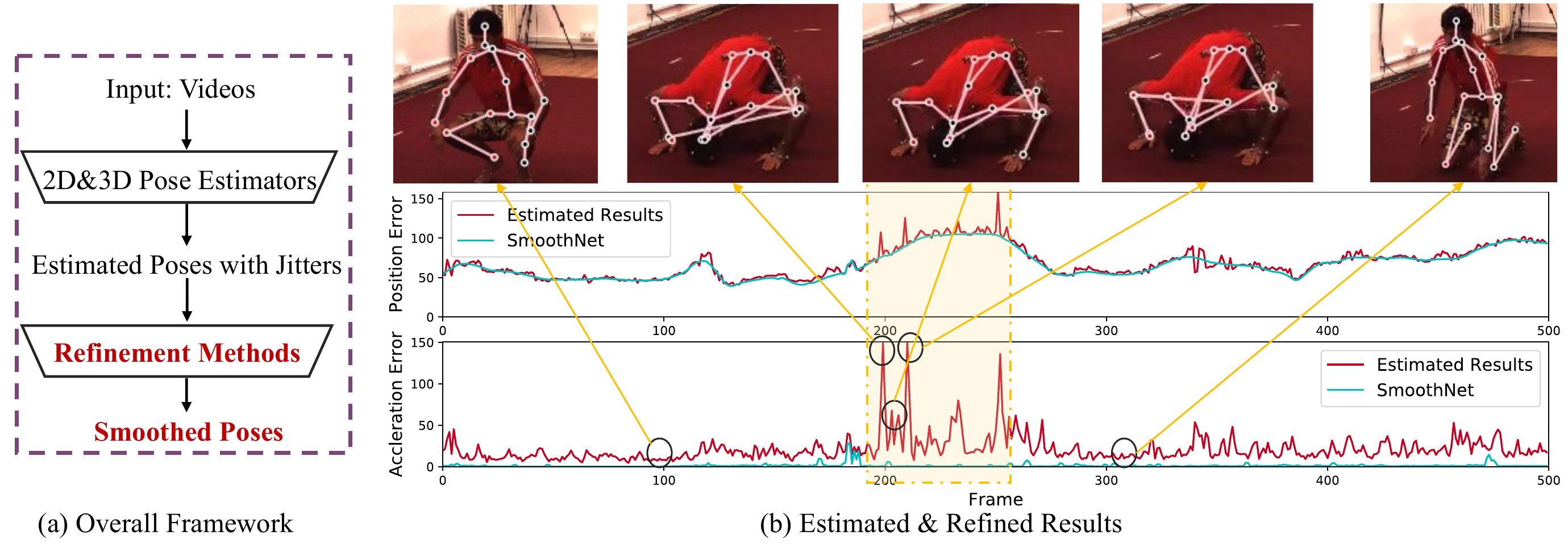}
\end{center}


\caption{\modelname~is a plug-and-play temporal-only refinement network for jitter mitigation. (a) shows the refinement flow. (b) demonstrates the estimated results of a state-of-the-art estimator RLE~\cite{li2021rle} and how \modelname~can improve the precision (upper curve) and smoothness (lower curve).}

\label{fig:mot2}
\end{figure*}

State-of-the-art end-to-end estimators such as~\cite{li2021rle,kocabas2020vibe} can output relatively accurate estimation results and small jitters for most frames (see Figure~\ref{fig:mot2}(b)). However, they tend to output large position errors when the video segments with rare/complex actions and these clips also suffer from significant jitters (e.g., from $200$ to $250$ frames in Figure~\ref{fig:mot2}(b)). 

The jitter problem in human pose estimation is hence highly unbalanced. 
Generally speaking, sudden jitters are easy to remove with low-pass filters~\cite{brownrigg1984weightedmedian,press1990savitzky,young1995gaus1d}. However, handling long-term jitters $\mathbf{J}$ is quite challenging because they usually entangle with ambiguous biased errors $\mathbf{S}$. 

\section{Related Work and Motivation}
\label{sec:relat}
\subsection{Spatio-Temporal Models for Smoothing}

Existing learning-based jitter mitigation solutions 
can be categorized into two types: end-to-end solutions and refinement networks after pose estimators. For the former category, various types of temporal models (e.g., gated recurrent units (GRUs)~\cite{kocabas2020vibe,choi2021tcmr,luo2020meva,zhou2022toch}, temporal convolutional networks (TCNs)~\cite{pavllo20193d,zeng2020srnet}, and Transformers~\cite{zheng20213d,wan2021encoder}) are used for temporal feature extraction. 
Other end-to-end solutions employ regularizers or loss functions to constrain the temporal consistency across successive frames~\cite{kanazawa2019learning,tripathi2020posenet3d,zhang2021learning,veges2020temporal,mehta2017vnect,mehta2020xnect}.
Recent pose refinement works~\cite{kim2021attention,veges2020temporal,jiang2021skeletor} take smoothness into consideration with spatial-temporal modeling.
%
Specifically, Jiang~\textit{et al}. \cite{jiang2021skeletor} designed a transformer-based network to smooth 3D poses in sign language recognition. 
Kim~\textit{et al}.~\cite{kim2021attention} propose a non-local attention mechanism with convolutions represented by \emph{quaternions}. 
Considering occlusions on multi-person scenes, Vege~\textit{et al}. \cite{veges2020temporal} conduct energy optimization with visibility scores to adaptively filter the keypoint trajectories. 

Without considering the highly-unbalanced nature of the jitter problem in pose estimation, the above solutions still cannot yield a smooth sequence of poses. There are mainly two reasons for such unsatisfactory performance. On the one hand, multiple joint positions largely deviate from the ground-truth for consecutive frames with long-term jitters, and the extracted spatial/temporal features themselves are untrustworthy, rendering less effective smoothing results. On the other hand, co-optimizing the jitter error $\mathbf{J}$ and the biased error $\mathbf{S}$ is challenging, and we name it the \emph{spatio-temporal optimization bottleneck.}\label{sec:rethink_st} 

We conduct comprehensive experiments on the popular 3D skeleton-based methods~\cite{martinez2017simple,pavllo20193d} and SMPL-based approaches~\cite{kolotouros2019spin,kocabas2020vibe} under single-frame, multi-frame, and smoothness loss settings. Due to space limitations, we put the results in the supplementary materials and summarize our key findings here: (i). compared to single-frame models, spatial-temporal models have better performance. However, the reduction in jitter errors $\mathbf{J}$ are still unsatisfactory (e.g., \emph{Accels} are reduced from $33$mm to $27$mm~\cite{kocabas2020vibe}); (ii). further adding an acceleration loss between consecutive frames or enhancing temporal modeling in the decoder design can benefit \emph{Accels} but harm MPJPEs (increase biased errors $\mathbf{S}$), due to the optimization bottleneck between per-frame precision and smoothness. 

With the above, a temporal-only pose smoothing solution is more promising for jitter mitigation.
Moreover, without using vastly different spatial information, such solutions have the potential to generalize across different datasets and motion modalities. 

\subsection{Low-Pass Filters for Smoothing}
%
Low-pass filters are general smoothing solutions, and they are used for pose refinement in the literature. 
For example, moving averages~\cite{hunter1986exponentially} that calculate the mean values over a specified period of time can be used to smooth sudden jitters.
%
Savitzky-Golay filter~\cite{press1990savitzky} uses a local polynomial least-squares function to fit the sequence within a given window size.
Gaussian filter~\cite{young1995gaus1d} modifies the input signal by convolution with a Gaussian function to obtain the minimum possible group delay.
Recently, a One-Euro filter was proposed in~\cite{casiez20121euro} for real-time jitter mitigation with an adaptive cutoff frequency. 

As a general temporal-only solution, low-pass filters can be applied to various pose refinement tasks without training. However, it inevitably faces the trade-off between jitters and lags, resulting in significant errors under long-term jitters. 

\noindent Motivated by the limitations of existing works, we propose a novel data-driven temporal-only refinement solution for 2D/3D human pose estimation tasks, as detailed in the following section. 

\section{Method}
\label{sec:method}

Instead of fusing spatial and temporal features for pose refinement, we explore long-range temporal receptive fields to capture robust temporal relations for mitigating large and long-term jitters. Specifically, the proposed \modelname~$g$ learns from the noisy estimated poses $\mathbf{\hat{Y}} \in \mathbb{R}^{L \times C}$ generated by any pose estimators $f$. 

The refinement function can be simply formulated as $\mathbf{\hat{G}} = g(\mathbf{\hat{Y}})$, where $\mathbf{\hat{G}}\in \mathbb{R}^{L \times C}$ is the smoothed poses.

\begin{figure}[ht]	
\centering
 	\subfigure[TCN block]
 	{
		\begin{minipage}[t]{0.3\linewidth}
			\centering      
			\includegraphics[width=1.5in]{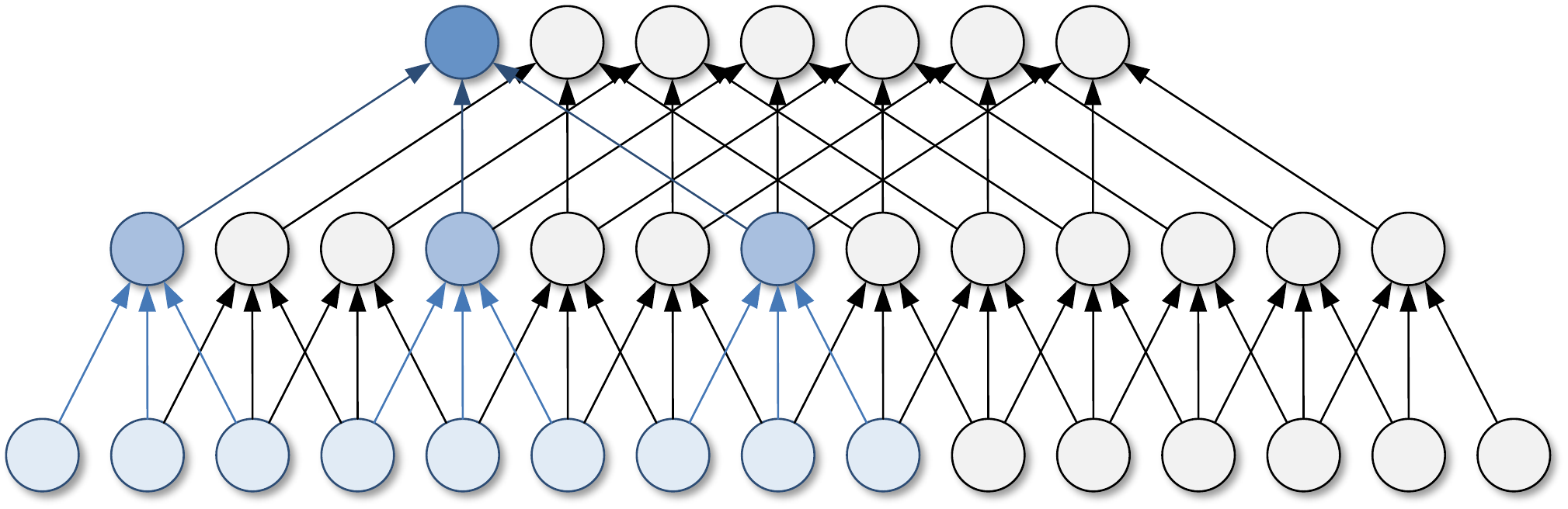}
		\end{minipage}
		\label{fig:framework_fc} 
	}
	\subfigure[Transformer block]
 	{
		\begin{minipage}[t]{0.3\linewidth}
			\centering      
			\includegraphics[width=1.5in]{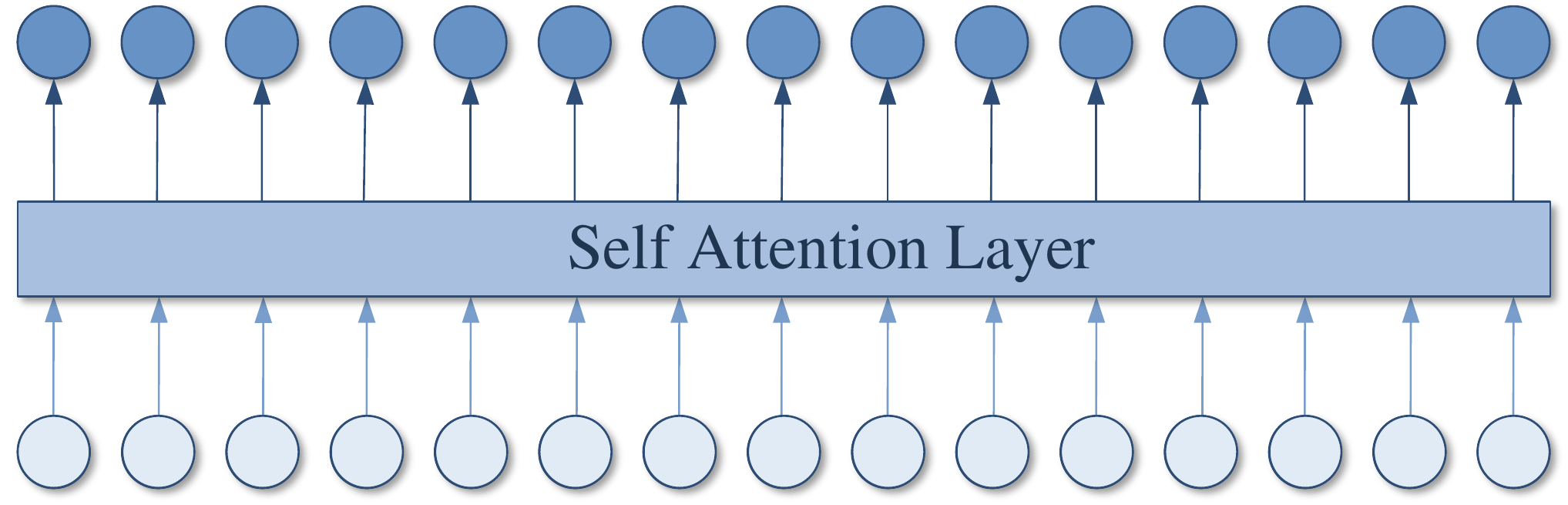}
		\end{minipage}
		\label{fig:framework_transformer} 
	}
	 \subfigure[\modelname~block] 
 	{
 		\begin{minipage}[t]{0.3\linewidth} 
 			\centering         
 			\includegraphics[width=1.5in]{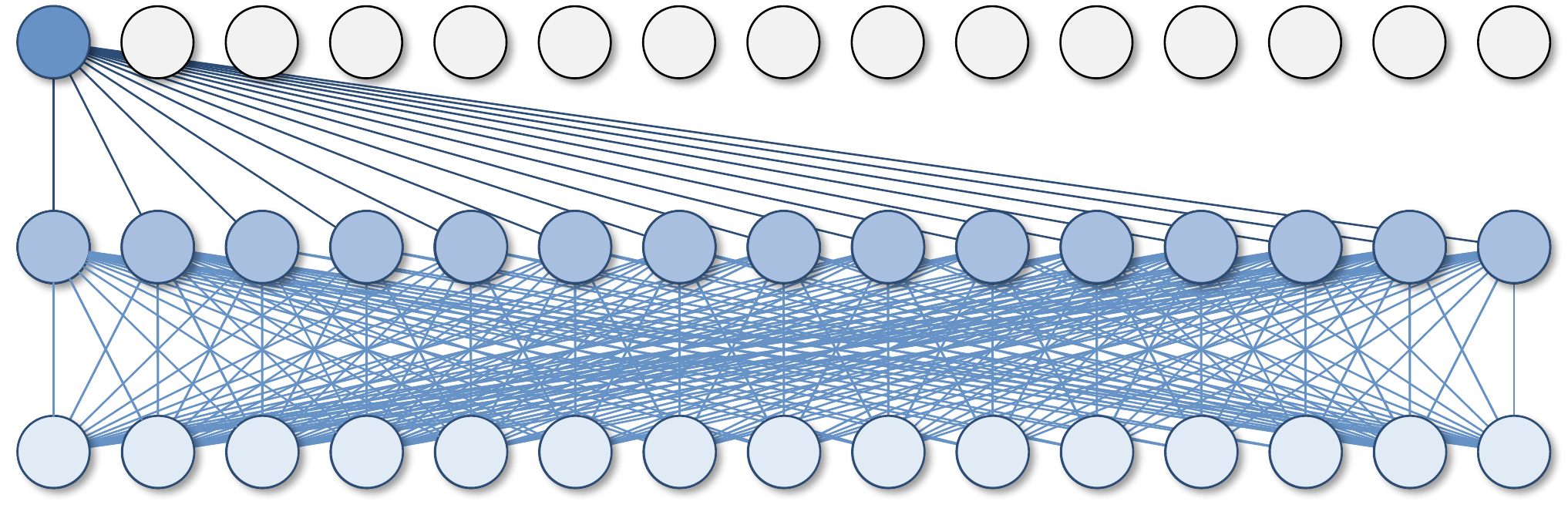}   
 		\end{minipage}
 		\label{fig:framework_tcn} 
 	}
\caption{Temporal relation extraction with (a) TCN, (b) Transformer, and (c) FCN. The input circles mean T frames information in one spatial dimension.}
\label{fig:compare_tcn_fc} 
\end{figure}

\subsection{Basic SmoothNet}
\label{sec:method_baseline}
Consider a fixed-length long sequence of the estimated joint positions, our objective is to capture temporal relations for smoothing. There are three popular temporal architectures that support long receptive fields without error accumulation, as illustrated in Fig.~\ref{fig:compare_tcn_fc}. 
Temporal convolutional networks (TCNs)~\cite{Bai2018tcn} conduct local convolutions (e.g., kernel size is 3) at each layer and employ dilation operations with multiple layers to enlarge the receptive field. 
In contrast, Transformers~\cite{vaswani2017attention} or fully-connected networks (FCNs) have global receptive fields at every layer, which can better tolerate long-term jitters than local convolutions in TCNs. While Transformers have become the \emph{de facto} sequence-to-sequence models in many application scenarios~\cite{zhou2021informer,zheng20213d,dosovitskiy2020image,so2019evolved}, we argue it is less applicable to our problem when compared to FCNs. In a Transformer model, the critical issue is to extract the semantic correlations between any two elements in a long sequence (e.g., words in texts or 2D patches in images) with self-attention mechanisms.
However, for pose refinement, we are more interested in modeling the continuity of motions on each joint (instead of point-wise correlations), which spans a continuous sequence of poses. 

Consequently, in this work, we propose to use FCNs as the backbone of our \modelname~design, which is position-aware and easy to train with abundant pose data from human motion videos. 
Additionally, according to the superposition of movements~\cite{fischman1984programming}, a movement can be decomposed as several movements performed independently. Based on this principle, each axis $i$ in channel $C$ can be processed independently.

\begin{figure}[h]
\begin{center}
\includegraphics[width=0.7\textwidth]{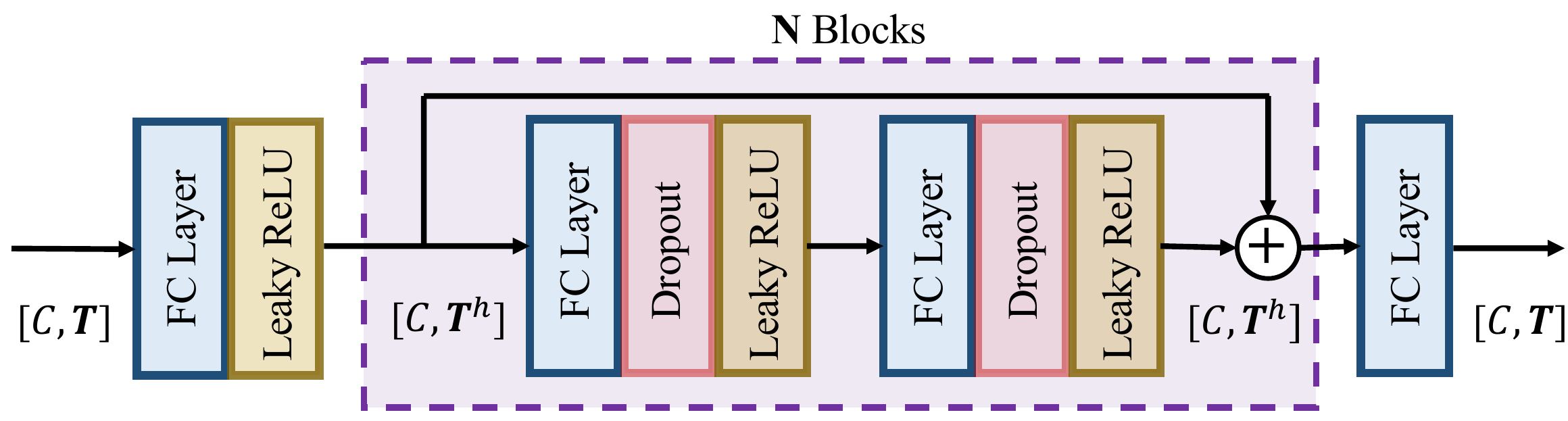}
\end{center}
\caption{A simple yet effective \modelname~design. }
\label{fig:m1}
\end{figure}

The proposed network is shown in Figure~\ref{fig:m1}, where we construct multiple FC layers with $N$ residual connected blocks along the temporal axis.
The computation of each layer can be formulated as follows.

\begin{equation}
    \small
    \hat{Y}_{i,t}^{l+1} = \sigma(\sum_{t=1}^T w_{t}^{l}*\hat{Y}_{i,t}^{l}+b^{l}),
    \label{eq:fc}
\end{equation}
where $w_{t}^{l}$ and $b^{l}$ are learnable weights and bias at the $t_{th}$ frame and they are shared among different $i_{th}$ axis, respectively.
$\sigma$ is the non-linear activation function (LeakyReLU is chosen by default).
To process $\mathbf{\hat{Y}}$ with \modelname, we adopt a sliding-window scheme similar to filters~\cite{press1990savitzky,young1995gaus1d,lee2001sliding}, where we first extract a chunk with size $T$, yield refined results thereon, and then move to the next chunk with a step size $s$ ($s \leq T$), preventing a loss of the last few frames.

\subsection{Motion-aware SmoothNet}
\label{sec:method_smooth}

\begin{figure}[h]
\begin{center}
\includegraphics[width=0.7\textwidth]{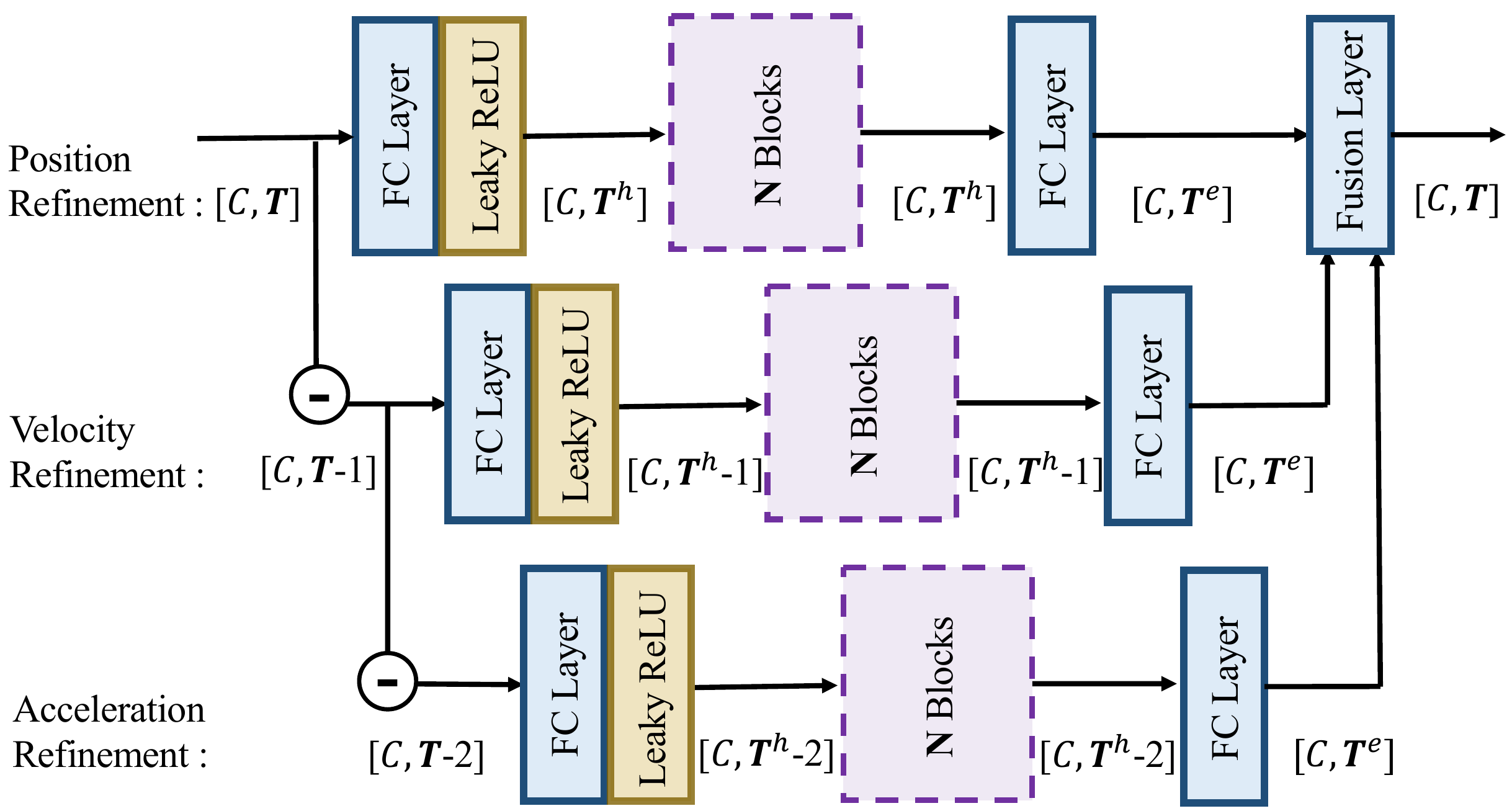}
\end{center}
\caption{The motion-aware \modelname~design. It explicitly models the velocity and acceleration with adjacent frames to achieve better pose refinement.}
\label{fig:m2}
\end{figure}

Our goal is to capture jitter patterns and reduce jitter errors $\mathbf{J}$, which mainly present as acceleration errors. It is straightforward to model acceleration explicitly in addition to position.
Accordingly, we further inject the movement function into our network, \textit{i.e.}, velocity and acceleration.
Given the prior with physical meaning, it is beneficial to leverage first-order and second-order motion information, making the learning process converge better and faster than the Basic \modelname.
Specifically, given the input $\mathbf{\hat{Y}}$, we first compute the velocity and acceleration (subtract by two consecutive frames) for each axis $i$, according to the Equation~\ref{eq:motion}.

\begin{equation}
    \small
    \hat{V}_{i, t} = \hat{Y}_{i,t} - \hat{Y}_{i,t-1}, \quad
    \hat{A}_{i, t} = \hat{V}_{i,t} - \hat{V}_{i,t-1}.
    \label{eq:motion}
\end{equation}
%
As shown in Figure~\ref{fig:m2}, the top branch is the baseline stream to refine noisy positions $\mathbf{\hat{Y}}$. The other two branches input the corresponding noisy velocity $\mathbf{\hat{V}}$ and acceleration $\mathbf{\hat{A}}$.
To capture the long-term temporal cue, we also employ Equation~\ref{eq:fc} to refine the velocity and acceleration. All branches consist of the same FC layers and blocks.
Then, we concatenate the top embedding of three branches to aggregate information from different order of motions and perform a linear fusion layer to obtain the final refined poses $\mathbf{\hat{G}}$.
Similar to the basic scheme in Section~\ref{sec:method_baseline}, this motion-aware scheme also works in a sliding-window manner to process the whole input sequence.

\subsection{Loss Function}
\label{sec:loss}
\modelname~aims to minimize both position errors and acceleration errors during training, and these objective functions are defined as follows.

\begin{equation}
\small
    L_{pose} = \frac{1}{T\times C}\sum_{t=0}^{T}\sum_{i=0}^{C} |\hat{G}_{i,t} - Y_{i,t}|,
    \label{eq:loss1}
\end{equation}
\begin{equation}
\small
    L_{acc} = \frac{1}{(T-2)\times C}\sum_{t=0}^{T}\sum_{i=0}^{C} |\hat{G}''_{i,t} - A_{i,t}|,
    \label{eq:loss2}
\end{equation}

where $\hat{G}''_{i,t}$ is the computed acceleration from predicted pose $\hat{G}_{i,t}$ and $A_{i,t}$ is the ground-truth acceleration.
We simply add $L_{pose}$ and $L_{acc}$ as our final target.

\section{Experiment}
\label{sec:exp}
We validate the effectiveness of the proposed~\modelname~and show quantitative results in the following sections.
Due to space limitations, we leave more analysis, discussions, and demos to the \emph{supplementary material}.
For more experimental details, please refer to the code.

\subsection{Experimental Settings}

\noindent \textbf{Backbones.}
We validate the generalization ability on both smoothness and precision of the proposed \modelname~covering three related tasks and several corresponding backbone models. 
For 2D pose estimation, we use Hourglass~\cite{newell2016stacked}, CPN~\cite{chen2018cascaded}, HRNet~\cite{sun2019hrnet} and RLE~\cite{li2021rle}; for 3D pose estimation, we implement FCN~\cite{martinez2017simple}, RLE~\cite{li2021rle}, TPoseNet~\cite{veges2020temporal} and VPose~\cite{pavllo20193d}; in terms of body recovery, we test on SPIN~\cite{kolotouros2019spin}, EFT~\cite{joo2020eft}, VIBE~\cite{kocabas2020vibe} and TCMR~\cite{choi2021tcmr}.

\noindent \textbf{Training sets.}
To prepare training data, we first save the outputs of existing methods, including estimated 2D positions, 3D positions, or SMPL parameters.
Then, we take these outputs as the inputs of \modelname~and use the corresponding ground-truth data as the supervision to train our model. 
%
In particular, we use the outputs of FCN on Human3.6M, SPIN on 3DPW \cite{von2018recovering}, and VIBE on AIST++ \cite{li2021aist} to train \modelname.
%

\noindent \textbf{Testing sets.}
We validate \modelname~on five dataset: Human3.6M~\cite{ionescu2013human3}, 3DPW \cite{von2018recovering}, MPI-INF-3DHP \cite{mehta2017monocular}, AIST++ \cite{li2021aist,tsuchida2019aist} and MuPoTS-3D~\cite{Mehta2018SingleShotM3} datasets. 

\noindent \textbf{Evaluation Metrics.}
To measure the jitter errors, we follow the related works~\cite{kanazawa2019learning,kocabas2020vibe,choi2021tcmr} to adopt \emph{Accel}. This is measured in \emph{mm/ frame$^{2}$} for 3D poses and \emph{pixel/ frame$^{2}$} for 2D poses. To evaluate the precision for each frame, besides \emph{MPJPE}, the \emph{Procrustes Analysis MPJPE (PA-MPJPE)} is another commonly used metric, where it removes effects on the inherent scale, rotation, and translation issues. For the 3D pose, the unit is \emph{mm}. For the 2D pose, we simply use \emph{pixel} in an image to validate the accurate localization precision.
%

\noindent \textbf{Implementation Details}
The basic \modelname~is an eight-layer model including the first layer, three cascaded blocks (N=3), and the last layer as a decoder. The motion-aware \modelname~contains three parallel branches with the first layer, one cascaded block, and the last layer for each branch. The input window size $T$ is $32$ and the moving step size $s$ is $1$. In addition, we use the sliding window average algorithm~\cite{lee2001sliding} based on smoothed results to avoid frame drop and reduce spikes. The parameters of \modelname~is $0.33$M, and the average inference time is less than 1.3k fps on a CPU and 46.8k on an A100-SXM4-40GB GPU.

\subsection{Comparison with Existing Solutions}
\label{sec:comparsion}
\subsubsection{Comparison with Filters}
\label{sec:single}

We compare \modelname~against three commonly used filters on the AIST++ dataset with pose estimator VIBE~\cite{kocabas2020vibe}. 
Experimental results are shown in Table~\ref{tab:filter_comp}. As can be observed, \modelname~achieves the best performance, and it reduces \emph{Accel} by $86.88$\% and \emph{MPJPE} by $8.82$\% compared to the original pose estimation results. 
Since we can easily trade off smoothness and lag in filter designs, there could be a large set of solutions with different \emph{Accel} and \emph{MPJPE} values. In this table, we present two possible solutions with the greedy search: one with comparable \emph{Accel} with \modelname~and the other with the minimum \emph{MPJPE}.

\begin{table}[h]
	\centering
    \caption{\textbf{Comparison \modelname~with widely-used filters on AIST++~\cite{li2021aist}.} The upper table with filters shows their lowest \emph{MPJPEs}, and the lower table is when their \emph{Accels} are comparable to ours. * means the inference speed is tested on a GPU.}
	{%
		\begin{tabular}{ll|cccc}
		    
			\specialrule{.1em}{.05em}{.05em}

			&Method& Accel & MPJPE& PA-MPJPE&Test FPS\\
			
			\midrule

            \parbox{3mm}{\multirow{10}{*}{\rotatebox[origin=c]{90}{Human Mesh Recovery}}} 
            &VIBE~\cite{kocabas2020vibe} &31.64&106.90&72.84&-\\
            \cmidrule{2-6}
            &w/ One-Euro~\cite{casiez20121euro} &10.82&108.55&74.67&2.31k\\
            &w/ Savitzky-Golay~\cite{press1990savitzky}&5.84 &105.80&72.15&31.22k\\
            &w/ Gaussian1d~\cite{young1995gaus1d}&4.95&103.42&71.11&37.45k\\
            \cmidrule{2-6}
            &w/ One-Euro~\cite{casiez20121euro} &4.67&135.71&103.22&2.43k\\
            &w/ Savitzky-Golay~\cite{press1990savitzky}&4.36 &118.25&85.39&30.19k\\
            &w/ Gaussian1d~\cite{young1995gaus1d}&4.47&105.71&71.49&38.21k\\
            \cmidrule{2-6}
            &\textbf{w/ Ours}&\textbf{4.15}&\textbf{97.47}&\textbf{69.67}&1.30k/\textbf{46.82k*}\\
        \midrule
        \end{tabular}%
	}

	\label{tab:filter_comp}
\end{table}

As a data-driven approach, \modelname~effectively learns the motion distribution of the complex movements in the dataset, resulting in much better \emph{MPJPEs} values, especially when \emph{Accel} is comparable. Among the three filters, the one-Euro filter shows inferior performance, and we attribute it to the real-time frame-by-frame jitter mitigation strategy used in it. 
Additionally, as \modelname~can benefit from GPU acceleration, it yields a much faster inference speed than filters (marked by *).
%

\begin{figure*}[h]	
\centering
    \subfigure[\emph{Accel} Distribution] 
	{
		\begin{minipage}[t]{0.45\linewidth}
			\centering      
			\includegraphics[width=2.2in]{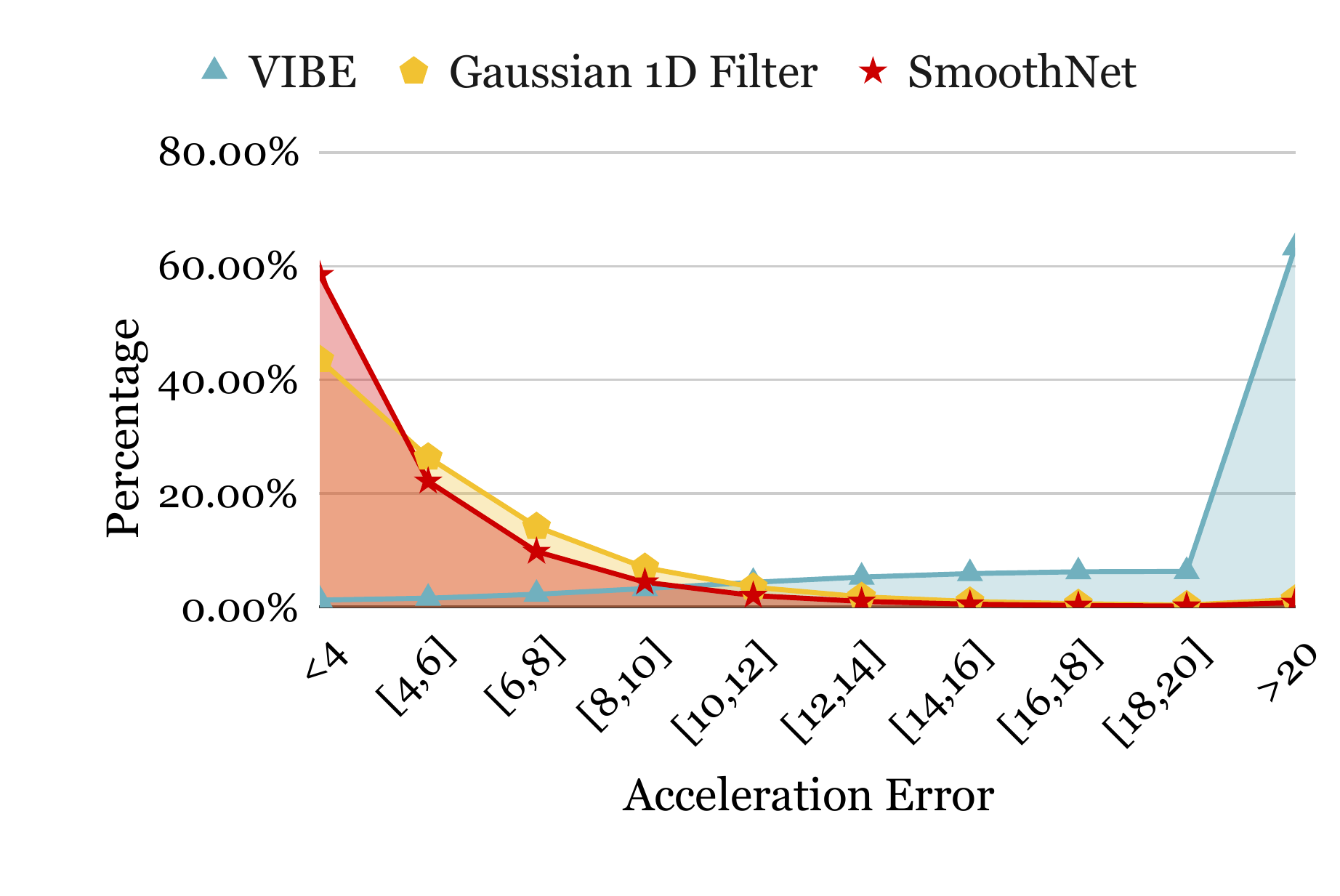}
		\end{minipage}
	}
    	\label{fig:acc_vibe}  
	\subfigure[\emph{MPJPE} Distribution] 
	{
		\begin{minipage}[t]{0.45\linewidth}
			\centering         
			\includegraphics[width=2.2in]{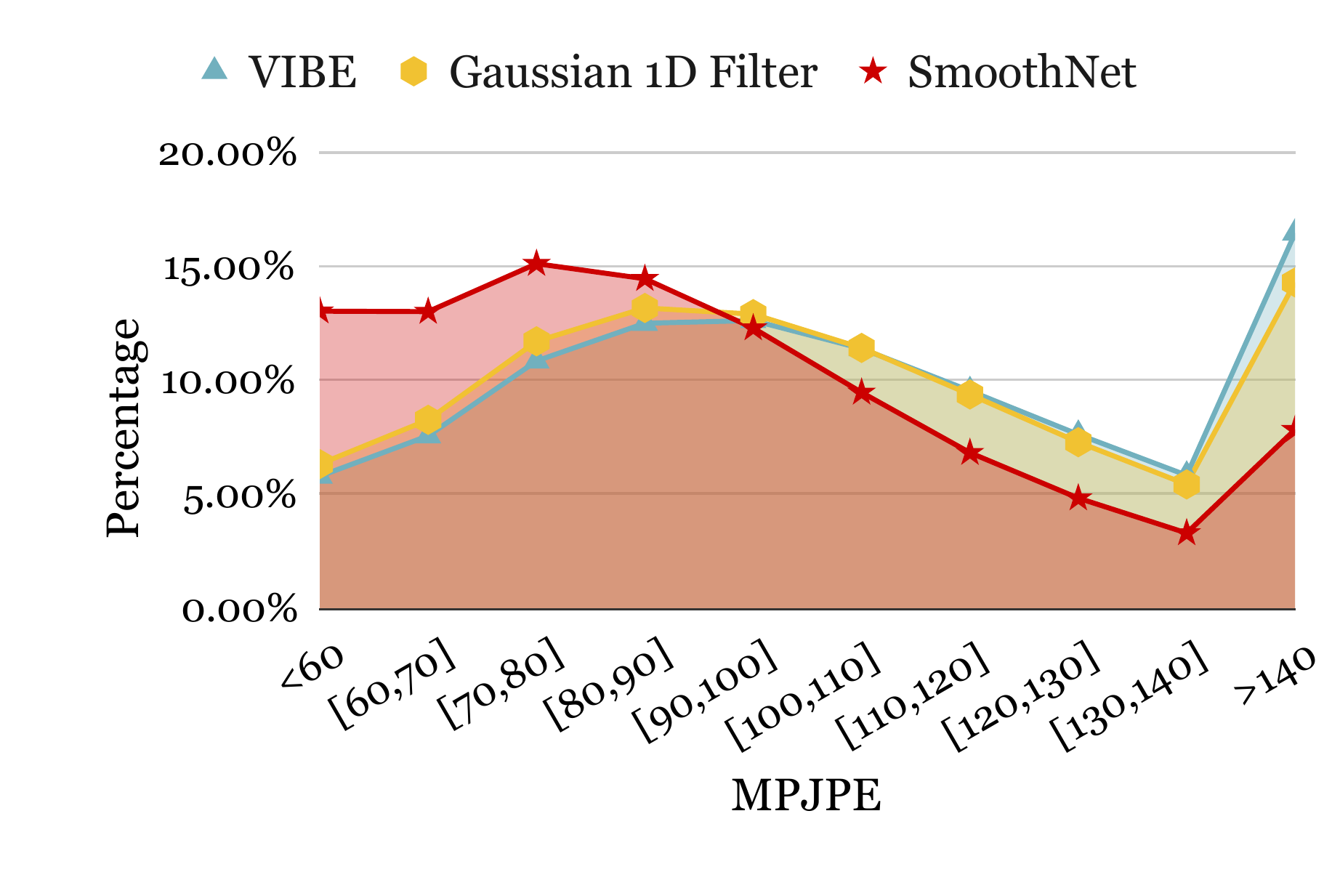}   
		\end{minipage}
	} 
    	\label{fig:mpjpe_vibe} 
\caption{Comparison of smoothness and precision distributions on AIST++. }
\label{fig:mpjpe_acc_smoothnet_vibe_gaussian} 
\end{figure*}

We further plot the \emph{MPJPE} and \emph{Accel} distribution of the original pose output from VIBE, VIBE with a Gaussian filter, and VIBE with \modelname~in Figure~\ref{fig:mpjpe_acc_smoothnet_vibe_gaussian}.
%
As can be observed, $98.7$\% of VIBE's original \emph{Accel} output falls above $4$ $mm/frame^2$. With Gaussian filter and \modelname, this percentage decreases to $56.5$\% and $41.6$\%, respectively.
As for MPJPEs, $5.78$\% of VIBE's outputs are smaller than $60$ $mm$ and $16.43$\% estimated poses are larger than $140$ $mm$. Gaussian filter increase the former proportion to $6.31$\% and decrease the latter proportion to $14.27$\%, improving precision slightly by removing some \emph{small} jitters and \emph{sudden} jitters. 
In contrast, \modelname~can increase the former percentage to $13.01$\% and decrease the latter percentage to $7.82$\% (a relative $45.2$\% reduction). 
We attribute the much higher performance of our solution to the fact that \modelname~can relieve large and long-term jitters effectively, thanks to its data-driven modeling of the smoothness characteristics in body movements.

\subsection{Refinement Results for Existing Methods}
As a plug-and-play network, \modelname~can be combined with any existing pose estimators. Here, we show the results on both skeleton-based methods and SMPL-based methods.

\begin{table}[h]
	\centering
	\vspace{-0.8cm}
    \small
    \caption{\textbf{Results of \modelname~attached to 2D and 3D pose estimators on Human3.6M dataset.} * is spatio-temporal backbones.}

	{%
		\begin{tabular}{ll|ccc|cc}
		    
            \midrule
			&Method & \emph{Accel} & \emph{MPJPE}& \emph{PA-MPJPE}&\emph{MPJPE-1\%}&\emph{Accel-1\%}\\
			\midrule
			\parbox{2.5mm}{\multirow{10}{*}{\rotatebox[origin=c]{90}{2D Pose Estimation}}}
            &Hourglass~\cite{newell2016stacked}&1.54& 9.42&7.64&55.81&2.71\\
            &\cellcolor{Gray}Hourglass w/ours&\cellcolor{Gray}\textbf{0.15}&\cellcolor{Gray}\textbf{9.25}&\cellcolor{Gray}\textbf{7.57}&\cellcolor{Gray}\textbf{55.50}&\cellcolor{Gray}\textbf{0.23}\\
            \cmidrule{2-7}
            &CPN~\cite{chen2018cascaded}&2.91&6.67&5.18&51.86&4.17\\
            &\cellcolor{Gray}CPN w/ours&\cellcolor{Gray}\textbf{0.14}&\cellcolor{Gray}\textbf{6.45}&\cellcolor{Gray}\textbf{4.96}&\cellcolor{Gray}\textbf{51.65}&\cellcolor{Gray}\textbf{0.22}\\
            \cmidrule{2-7}
            &HRNet~\cite{sun2019hrnet}&1.01&4.59&4.19&18.16&3.55\\
            &\cellcolor{Gray}HRNet w/ours&\cellcolor{Gray}\textbf{0.13}&\cellcolor{Gray}\textbf{4.54}&\cellcolor{Gray}\textbf{4.13}&\cellcolor{Gray}\textbf{16.98}&\cellcolor{Gray}\textbf{0.26}\\
            \cmidrule{2-7}
            &RLE ~\cite{li2021rle}&0.90&5.14&4.82&16.67&2.28\\
            &\cellcolor{Gray}RLE w/ours&\cellcolor{Gray}\textbf{0.13}&\cellcolor{Gray}\textbf{5.21}&\cellcolor{Gray}\textbf{4.78}&\cellcolor{Gray}\textbf{16.16}&\cellcolor{Gray}\textbf{0.19}\\
			\midrule
		    \parbox{0.1mm}{\multirow{12}{*}{\rotatebox[origin=c]{90}{3D Pose Estimation}}} 
            &FCN~\cite{martinez2017simple}&19.17&54.55&42.20&161.00&40.03\\
            &\cellcolor{Gray}FCN w/ours&\cellcolor{Gray}\textbf{1.03}&\cellcolor{Gray}\textbf{52.72}&\cellcolor{Gray}\textbf{40.92}&\cellcolor{Gray}\textbf{151.08}{\color{Red}$\downarrow_{6.2\%}$}&\cellcolor{Gray}\textbf{1.52}{\color{Red}$\downarrow_{96.2\%}$}\\
            \cmidrule{2-7}
            &RLE~\cite{li2021rle} &7.75&48.87&38.63&139.04&16.54\\
            &\cellcolor{Gray}RLE w/ours&\cellcolor{Gray}\textbf{0.90}&\cellcolor{Gray}\textbf{48.27}&\cellcolor{Gray}\textbf{38.13}&\cellcolor{Gray}\textbf{136.70}{\color{Red}$\downarrow_{1.7\%}$}&\cellcolor{Gray}\textbf{1.01}{\color{Red}$\downarrow_{93.9\%}$}\\
            \cmidrule{2-7}
            &VPose~\cite{pavllo20193d} (T=27)* & 3.53&50.13&39.13&153.87&7.95\\
            &\cellcolor{Gray}VPose (T=27)* w/ours&\cellcolor{Gray}\textbf{0.88}&\cellcolor{Gray}\textbf{50.04}&\cellcolor{Gray}\textbf{39.04}&\cellcolor{Gray}\textbf{153.29}{\color{Red}$\downarrow_{3.8\%}$}&\cellcolor{Gray}\textbf{0.94}{\color{Red}$\downarrow_{88.2\%}$}\\
            \cmidrule{2-7}
            &VPose (T=81)*&3.06 &48.97&38.27&149.97&6.52\\
            &\cellcolor{Gray}VPose (T=81)* w/ours&\cellcolor{Gray}\textbf{0.87}&\cellcolor{Gray}\textbf{48.89}&\cellcolor{Gray}\textbf{38.21}&\cellcolor{Gray}\textbf{149.57}{\color{Red}$\downarrow_{0.3\%}$}&\cellcolor{Gray}\textbf{0.85}{\color{Red}$\downarrow_{87.0\%}$}\\
            \cmidrule{2-7}
            &VPose (T=243)*&2.82 &48.11&37.71&150.25&6.01\\
            &\cellcolor{Gray}VPose (T=243)* w/ours&\cellcolor{Gray}\textbf{0.87}&\cellcolor{Gray}\textbf{48.05}&\cellcolor{Gray}\textbf{37.66}&\cellcolor{Gray}\textbf{149.88}{\color{Red}$\downarrow_{0.2\%}$}&\cellcolor{Gray}\textbf{0.83}{\color{Red}$\downarrow_{86.2\%}$}\\

        \midrule
        \end{tabular}%
	}
    \begin{tablenotes} 
    \tiny
		\item  All estimation results are re-implemented or tested by us for fair comparisons.
     \end{tablenotes}
	\label{tab:hm_results}
\end{table}

\subsubsection{2D and 3D Pose Estimation}
\label{sec:comb_skeleton}

In Table~\ref{tab:hm_results}, we compare the results of skeleton-based methods on the Human3.6M dataset. The \emph{Accel} of all the backbones followed by our pose refinement method is significantly reduced, and \emph{MPJPE} is also reduced to some extent. 
Specifically, \emph{Accel} and \emph{MPJPE} are reduced to a greater extent for the single-frame networks.
Also, we observe that the refined \emph{Accel} is similar with different backbones, indicating that \modelname~can effectively remove different kinds of jitters in pose estimation. 
Since \modelname~is only trained with FCN-Human3.6M, the improvements on FCN~\cite{martinez2017simple} is larger than other backbones with $94.6$\%, $3.4$\% and $3.0$\% reduction in \emph{Accel}, \emph{MPJPE} and \emph{PA-MPJPE}, respectively.
To explore the impact on significant biased errors and long-term jitters, we further calculate the largest $1$\% of \emph{MPJPE} (\emph{MPJPE-1\%}) and their corresponding \emph{Accel} (\emph{Accel-1\%}) as the worst $1$\% estimated poses for each backbone. On average, the estimated \emph{Accel} on these poses are decreased by about $90$\%. In particular, we could achieve an $6.2$\% improvement on the trained backbone FCN with \emph{MPJPE} reduced from $161.00mm$ to $151.08mm$. This is because significant position errors are usually accompanied by long-term and large jitters, and \modelname can reduce them as a side-effect during smoothing. Moreover, the above results across backbones also validate the generalization capability of~\modelname.

\subsubsection{Human Mesh Recovery}
\label{sec:comb_smpl}

In Table~\ref{tab:smpl_p}, we give results of SMPL-based methods for body recovery on 3DPW\cite{von2018recovering}, MPI-INF-3DHP \cite{mehta2017monocular}, and Human3.6M dataset~\cite{ionescu2013human3}. 
\modelname~is trained with the pose outputs from SPIN~\cite{kolotouros2019spin}. We test its performance across multiple backbone networks.

\begin{table*}[h]
	\centering
	\caption{\textbf{Results of \modelname~attached to human mesh recovery models} on 3DPW~\cite{von2018recovering}, MPI-INF-3DHP~\cite{mehta2017monocular}, and Human3.6M~\cite{ionescu2013human3} dataset. * is spatio-temporal backbones.}

	\resizebox{\textwidth}{!}{%
		\begin{tabular}{l|ccc|ccc|ccc}
		    
			\specialrule{.1em}{.05em}{.05em}

			\multirow{2}{*}{Method}& \multicolumn{3}{c}{3DPW} & \multicolumn{3}{c}{MPI-INF-3DHP} & \multicolumn{3}{c}{Human3.6M} \\
			\cmidrule(lr){2-4} \cmidrule(lr){5-7} \cmidrule(lr){8-10}
		& \emph{Accel}  & \emph{MPJPE}&\emph{PA-MPJPE}  & \emph{Accel}  & \emph{MPJPE}  & \emph{PA-MPJPE} & \emph{Accel}& \emph{MPJPE} &\emph{PA-MPJPE}  \\
			
			\midrule

			SPIN~\cite{kolotouros2019spin} & 30.8  & 87.6 &  53.3&28.5 & 100.2 & 61.4 & 18.6 & 68.5 & 46.5 \\
			\cellcolor{Gray}SPIN w/ours& \cellcolor{Gray} \textbf{5.5}& \cellcolor{Gray} \textbf{86.7}& \cellcolor{Gray} \textbf{52.7}& \cellcolor{Gray}\textbf{6.5} & \cellcolor{Gray}\textbf{92.9} & \cellcolor{Gray}\textbf{60.2}& \cellcolor{Gray} \textbf{2.8}& \cellcolor{Gray} \textbf{67.5}& \cellcolor{Gray}\textbf{46.3} \\ 

			\midrule
			
			VIBE*~\cite{kocabas2020vibe}&23.2  &83.0 & 52.0& 22.3 & 91.9 & 58.9 & 15.8 &78.1& 53.7  \\
			 \cellcolor{Gray}VIBE* w/ours& \cellcolor{Gray}\textbf{6.0}& \cellcolor{Gray}\textbf{81.5} & \cellcolor{Gray} \textbf{51.7}& \cellcolor{Gray}\textbf{6.5} & \cellcolor{Gray}\textbf{87.6} & \cellcolor{Gray}\textbf{58.8}& \cellcolor{Gray}\textbf{2.9} & \cellcolor{Gray}\textbf{77.2} & \cellcolor{Gray}\textbf{53.4} \\
            \midrule
            TCMR*~\cite{choi2021tcmr} & 6.8 & \textbf{86.5}&  \textbf{52.7} & 8.0 & 92.6 & \textbf{58.2} & 3.8& \textbf{73.6}  & \textbf{52.0}  \\ 
			TCMR w/MEVA*~\cite{luo2020meva}& 6.2 &  88.7& 55.0& -& -& -&  3.1&  77.2& 55.4 \\
			\cellcolor{Gray}TCMR* w/ours& \cellcolor{Gray}\textbf{6.0} & \cellcolor{Gray} \textbf{86.5}& \cellcolor{Gray}53.0 & \cellcolor{Gray}\textbf{6.5} & \cellcolor{Gray} \textbf{88.9}& \cellcolor{Gray}58.9& \cellcolor{Gray} \textbf{2.8}& \cellcolor{Gray}73.9 & \cellcolor{Gray}52.1 \\
            \specialrule{.1em}{.05em}{.05em}
		\end{tabular}%
	}
	\begin{tablenotes} 
    \tiny
		\item  All estimation results are re-implemented or tested by us for fair comparisons.
    \end{tablenotes}
	\label{tab:smpl_p}
\end{table*}

Overall, our method has a consistent improvement in smoothness and precision. 
Specifically, \modelname~can reduce \emph{Accel} on SPIN and VIBE by a large margin. Compared to the original estimated poses from SPIN, our method improves by about $82.1$\% and $1.0$\% on \emph{Accel} and \emph{MPJPE}, respectively.
For the TCMR backbone, since it has used some smoothing strategies in its models, their original \emph{Accel} is relatively small. However, the first and last few frames could not be smoothed out with their method. Our model can relieve such jitters and further enhance its performance. Moreover, we add the post-processing slerp filter to minimize Euclidean distance on quaternion from MEVA~\cite{luo2020meva} on top of TCMR backbone. The filter can improve \emph{Accel}, but causes over-smoothness, leading to higher position errors.
\subsection{Ablation Study}
\label{sec:ablation}

\noindent\textbf{Comparisons on Temporal Models.}
To further validate the capability of the proposed FCN-based temporal model \modelname, we compare it with (i). traditional Gaussian1d filter; (ii). temporal convolutional networks~\cite{Bai2018tcn} with a small kernel size (here is $3$) in each layer, with $6$, $8$, $10$ layers to obtain $27$, $81$, and $243$ final receptive fields, respectively; (iii). self-attention-based Transformer (Trans.); (iv). TCN (81)$\times$ with overlapped sliding window scheme to enhance the output quality.
%
Same as the Section~\ref{sec:single}, we use inputs from VIBE-AIST++.

\begin{table}[t]
	\centering
    \caption{\textbf{Comparison results with different temporal models on VIBE-AIST++.} $\times$ is to use overlapped sliding-window scheme, which is used in \modelname~by default. Ours$\star$ is the same model with a non-overlapping sliding window. }
    \resizebox{\textwidth}{10mm}
	{%
		\begin{tabular}{l|cccccc|cc}

			\specialrule{.1em}{.05em}{.05em}
			
			Method&Gaussian1d&TCN(27) & TCN(81)& TCN(81)$\times$&TCN(243) &Trans.$\times$&Ours$\star$&Ours$\times$ \\
			\midrule
			\emph{Accel}&4.95&14.46&11.84&8.71&10.07&6.15&\underline{5.45}&\textbf{ 4.15}\\
			\emph{MPJPE}&103.42&103.53&101.17&99.54&99.76&99.30&\underline{98.34}&\textbf{97.47}\\
			\emph{PA-MPJPE}&71.11&72.99&72.30&71.80&71.92&71.89&\underline{71.02}&\textbf{69.67}\\

        \midrule
        \end{tabular}%
	}
	\label{tab:conv}
\end{table}

Results are shown in Table~\ref{tab:conv}, which indicates (i). the performance of TCN improves with increased receptive fields;
(ii). the \emph{Accel} of TCNs are worse than that of the filter~\cite{young1995gaus1d}, implying local aggregation of noisy poses with the shared kernels cannot handle large and long-term jitters well; 
(iii) the \emph{MPJPE} of TCNs and Transformers are lower than that of the filter, indicating learning-based methods can further reduce biased errors $S$ with learning the noisy pose prior; 
(iv) Transformer achieves a good balance between \emph{Accel} and \emph{MPJPE} with the global receptive field at each layer, but not as good as \modelname. We attribute it to the unnecessary self-attention operations for the pose refinement task, which is no guarantee to model the smoothness pattern well.
Lastly, our method show superiority in all metrics even without overlapped sliding window scheme. Note that the sliding window scheme can relieve the spikes at the junction of two sliding windows, especially when MPJPE is huge.

\begin{figure}[t]
\begin{center}
\includegraphics[width=0.6\textwidth]{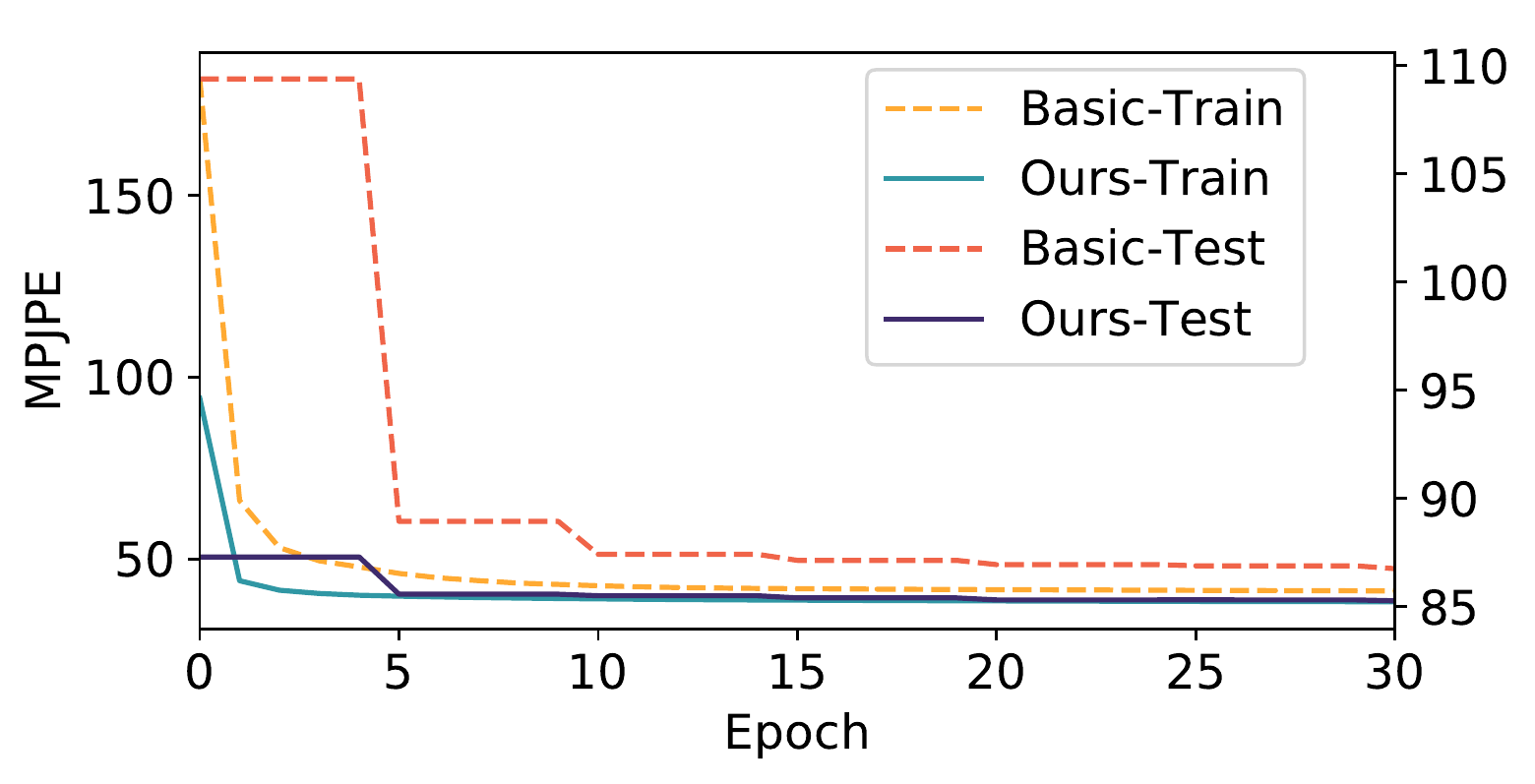}
\end{center}
\caption{Impact of model designs from the training and testing precision curves. }
\label{fig:model}
\end{figure}
\noindent\textbf{Comparison between the Two Proposed Models.}
To capture the long-range temporal relations from noisy estimated pose sequences, we first propose a simple model with the residual fully connected network on temporal dimension, named \emph{basic \modelname}. To further improve performance, we design a motion-aware temporal network as the \modelname~in Sec.~\ref{sec:method_smooth}. Figure~\ref{fig:model} illustrates the training and testing precision curves of these two models on 3DPW. We can observe that (i) \emph{basic} model tends to somewhat overfit; (ii) \modelname~fits better and obtain slightly lower position errors. In comprehensive studies, we summarize the motion-aware \modelname~can fit better than the basic one, while the basic one can obtain impressive results with its simple design.

\noindent\textbf{Impact of Window Size.}
The window size $W$ will largely impact of smoothness from previous sliding-window-based methods~\cite{kim2021attention,choi2021tcmr,kocabas2020vibe,press1990savitzky}. We demonstrate the effects on different window sizes from $2$ to $256$ frames in Table~\ref{tab:window}. As the window size becomes longer, the \emph{Accel} decreases consistently, but the \emph{MPJPE} and \emph{PA-MPJPE} initially decrease, then begin to increase slightly, indicating that when the size exceeds $64$ frames, the results of the three metrics tend to be saturated. Therefore, $64$ frames can be suitable to balance the smoothness and precision.

\begin{table}[h]
	\centering
    \caption{\textbf{Impact of window size $W$ on VIBE-AIST++~\cite{li2021aist}.} }
 
	{%
		\begin{tabular}{l|ccccccccc}
		\midrule
	    	$W$&VIBE&2&8&16&32&64&128&256\\
	    	\midrule
		    \emph{Accel}&31.63&17.89&5.76&4.54&4.15&4.07&4.04&\textbf{4.03}\\
		    \emph{MPJPE}& 106.90&102.57&99.98&98.62&97.47&97.06&\textbf{93.20}&94.89\\
		    \emph{PA-MPJPE}&72.84&71.48&70.51&69.85&\textbf{69.67}&69.89&70.57&71.52\\
        \midrule
        \end{tabular}%
	}
	\label{tab:window}
\end{table}

\section{Conclusion}
In this work, we propose \modelname, a simple yet effective pose refinement network to improve the temporal smoothness and per-frame precision of existing pose/body estimators. Compared to existing solutions, \modelname~can deal with long-term significant jitters, which often occur with rare/complex poses, as verified with comprehensive experiments on a large number of backbone networks, commonly modalities and datasets.
 
\noindent \textbf{Broader Impact:}
\modelname~is a temporal-only model targeting at removing various jitters, which takes advantage of the continuity of human motion, and generalizes well across backbones, modalities, and even datasets. Accordingly, this idea could be applied to other related tasks, such as whole-body estimation, pose tracking, and multi-object tracking, to further improve their smoothness and precision. Moreover, \modelname~could potentially provide a smoothness prior over human motion, which is complementary to pose prior VPoser~\cite{choutas2020monocular} and motion prior MPoser~\cite{kocabas2020vibe}.

\noindent \textbf{Limitation and Future Work:} 
\modelname~is a sliding-window-based model, which limits its use in real-time systems since we can not aggregate future poses to refine the historical poses. A real-time and accurate refinement model will be beneficial for online applications. We leave them for future work.

\noindent \textbf{Acknowledgement.} This work is supported in part by Shenzhen-Hong Kong-Macau Science and Technology Program (Category C) of Shenzhen Science Technology and Innovation Commission under Grant No. SGDX2020110309500101.

\clearpage
%
%

\bibliographystyle{splncs04}

\end{document}


\pagestyle{headings}
\mainmatter
\def\ECCVSubNumber{1848}  

\title{---Supplementary Materials---\\SmoothNet: A Plug-and-Play Network for Refining Human Poses in Videos} 

\titlerunning{SmoothNet}
%
\author{Ailing Zeng$^{1}$ \and
Lei Yang$^{2}$ \and
Xuan Ju$^{1}$ \and
Jiefeng Li$^{3}$ \and
Jianyi Wang$^{4}$ \and
Qiang Xu$^{1}$}
\authorrunning{A. Zeng et al.}
%
\institute{$^{1}$The Chinese University of Hong Kong, $^{2}$
Sensetime Group Ltd., \\$^{3}$Shanghai Jiao Tong University, $^{4}$Nanyang Technological University \\
\email{\{alzeng, qxu\}@cse.cuhk.edu.hk}}
\maketitle

In this supplementary material, we present additional experimental details about dataset descriptions, implementation details in Sec.~\ref{sec:supp_exp}. In Sec.~\ref{sec:supp_exp_analyses}, we show more experimental analyses on existing Spatio-temporal models, comparison with filters on 2D/3D pose estimation, analysis of additional metrics, comparison with learnable RefineNet, and smoothness on synthetic data. Moreover, we conduct more ablation studies on the effect of the loss function, motion modalities, normalization strategies
which are not shown in the main paper due to the space limitation. Lastly, in Sec.~\ref{sec:supp_viz}, we visualize qualitative results to verify the effectiveness and necessity of \modelname. For more visualization, please refer to our website\footnote{Website: \url{https://ailingzeng.site/smoothnet}}
 
\section{Experimental Details}
\label{sec:supp_exp}
\subsection{Dataset Description}
\noindent -- \textit{Human3.6M}~\cite{ionescu2013human3} consists of $3.6$ million frames' $50$ fps videos with $15$ actions from $4$ camera viewpoints. 3D human joint positions are captured accurately from a high-speed motion capture system. We can use the camera intrinsic parameters to calculate their accurate 2D joint positions. Following previous works~\cite{zeng2020srnet,martinez2017simple,pavllo20193d}, we adopt the standard cross-subject protocol with $5$ subjects (S1, S5, S6, S7, S8) as the training set and another $2$ subjects (S9, S11) as the testing set.

\noindent -- \textit{3DPW} \cite{von2018recovering} an in-the-wild dataset consisting of more than $51,000$ frames' accurate 3D poses in challenging sequences with 30 fps. It is usually used to validate the effectiveness of model-based methods~\cite{kolotouros2019spin,kocabas2020vibe,choi2021tcmr}.

\noindent -- \textit{AIST++} \cite{li2021aist} is a challenging dataset that comes from the AIST Dance Video DB~\cite{tsuchida2019aist}. It contains $1,408$ 3D human dance motion sequences with $60$ fps, providing 3D human keypoint annotations and camera parameters for $10.1$M images, covering $30$ different subjects in $9$ views. We follow the original settings to split the training and testing sets.

\noindent -- \textit{MPI-INF-3DHP} \cite{mehta2017monocular} contains both constrained indoor scenes and complex outdoor scenes, covering a great diversity of poses and actions. It is usually used to verify the generalization ability of the proposed methods. We use this dataset as the testing set.

\noindent -- \textit{MuPoTS-3D}~\cite{Mehta2018SingleShotM3} is a testing set for multi-person 3D human pose, containing $20$ indoor and outdoor video sequences. We also use it as the testing set.

\subsection{Implementation Details}
For data preprocessing, we normalize 2D positions into [$-1$, $1$] by the width and length of the videos, and we use root-relative 3D positions with the unit of meter, where they can range in [$-1$, $1$]. For SMPL estimation, we use the original 6D rotation matrix without any normalization.

For the usage of motion modalities, in the training stage, we use 3D positions to train \modelname~by default. 
Because \modelname~shares its weights as well as biases among different spatial dimensions, it can be used directly across different motion modalities. In the inference stage, we can use the trained model to test different motion modalities.
If the number of skeleton points is $N$, the outputs of 2D (C = $2*N$) and 3D (C = $3*N$) pose estimation are a series of 2D and 3D positions. The outputs of mesh recovery are the pose parameters as 6D rotation matrix~\cite{Zhou2019OnTC} (C = $6*N$), $10$ shape parameters and $3$ camera parameters. Different datasets have different $N$ (e.g. $N$ is $17$ in Human3.6M, MPI-INF-3DHP and MuPoTS-3D, $N$ is $24$ in 3DPW and AIST++).

For the AIST++ dataset~\cite{li2021aist}, we find that some inaccurate fitting from SMPLify causes misleading supervision in 6D rotation and high errors because of lacking enough keypoints as constraints. Thus, we simply threw away the test videos with \emph{MPJPE}s (computed by the estimated results of VIBE and the given ground truth) bigger than $170$mm.

For training details, the initial learning rate is $0.001$, and it decays exponentially with the rate of $0.95$. We train the proposed model for $70$ epochs using Adam optimizer. The mini-batch size is $128$. Our experiments can be conducted on a GPU with an NVIDIA GTX 1080 Ti. 

For hyperparameters of filters, in the lower part of Table 1 in the main paper, we set the window size of the Savitzky-Golay filter as $257$ and the polyorder (order of the polynomial used to fit the samples) as $2$ to obtain the comparable Acceleration errors with us. For the Gaussian1d filter, we set the sigma (standard deviation for Gaussian kernel) as $4$ and window size as $129$. For the One-Euro filter, the cutoff (the minimum cutoff frequency) is $1e^{-4}$, and the lag value (the speed coefficient) is $0.7$. Meanwhile, in the upper part of Table 1, to obtain comparable \emph{MPJPE}s, we set $31$ as the window size with the polyorder as $2$ for the Savitzky-Golay filter. We apply $31$ as the window size with sigma as $3$ for the Gaussian1d filter and modify the cutoff to $0.04$ for the One-Euro filter.
In addition, we follow the common tools to implement \textit{One-Euro}\footnote{\scriptsize \url{https://github.com/mkocabas/VIBE/blob/master/lib/utils/one\_euro\_filter.py}}, \textit{Savitzky-Golay}\footnote{\scriptsize\url{https://docs.scipy.org/doc/scipy/reference/generated/scipy.signal.savgol\_filter.html}} and \textit{Gaussian1d filters}\footnote{\scriptsize\url{https://docs.scipy.org/doc/scipy/reference/generated/scipy.ndimage.gaussian\_filter1d.html}}.

\section{Experimental Analyses}
\label{sec:supp_exp_analyses}

\begin{figure*}[hb]	
	\subfigure[3D Skeleton-based Methods~\cite{martinez2017simple,pavllo20193d}] 
	{
		\begin{minipage}[t]{0.47\linewidth}
			\centering         
			\includegraphics[width=2.45in]{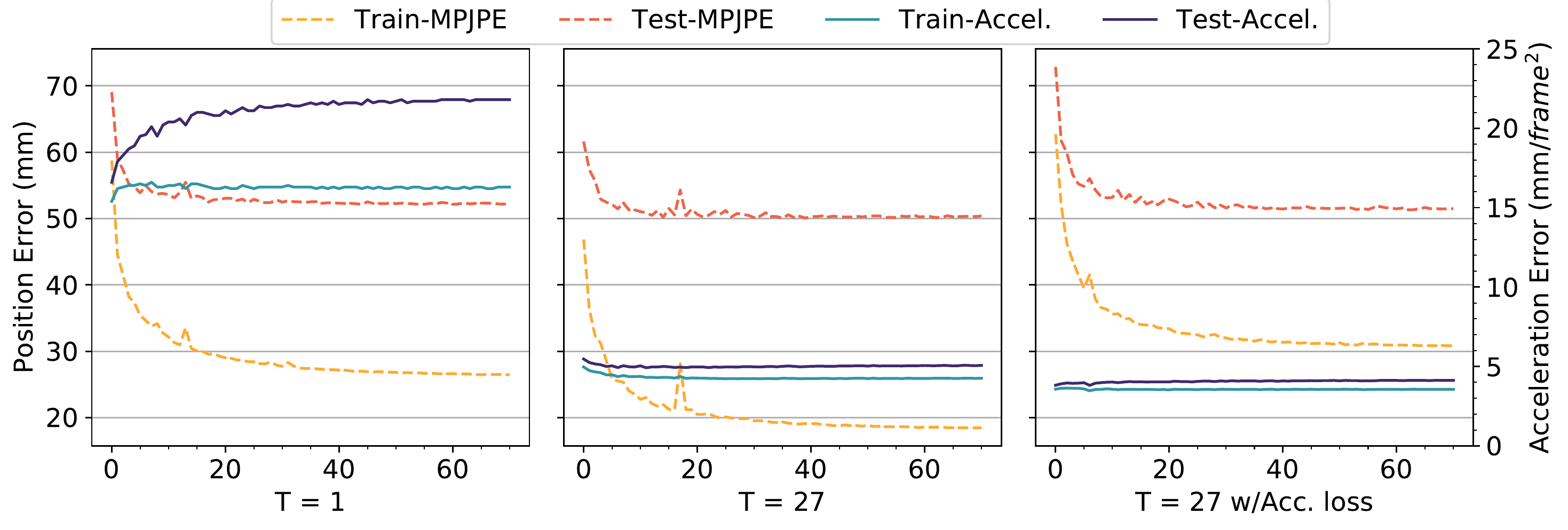}   
		\end{minipage}
	} 
    	\label{fig:ske_result} 
	\subfigure[SMPL-based Methods~\cite{kocabas2020vibe}] 
	{
		\begin{minipage}[t]{0.47\linewidth}
			\centering      
			\includegraphics[width=2.45in]{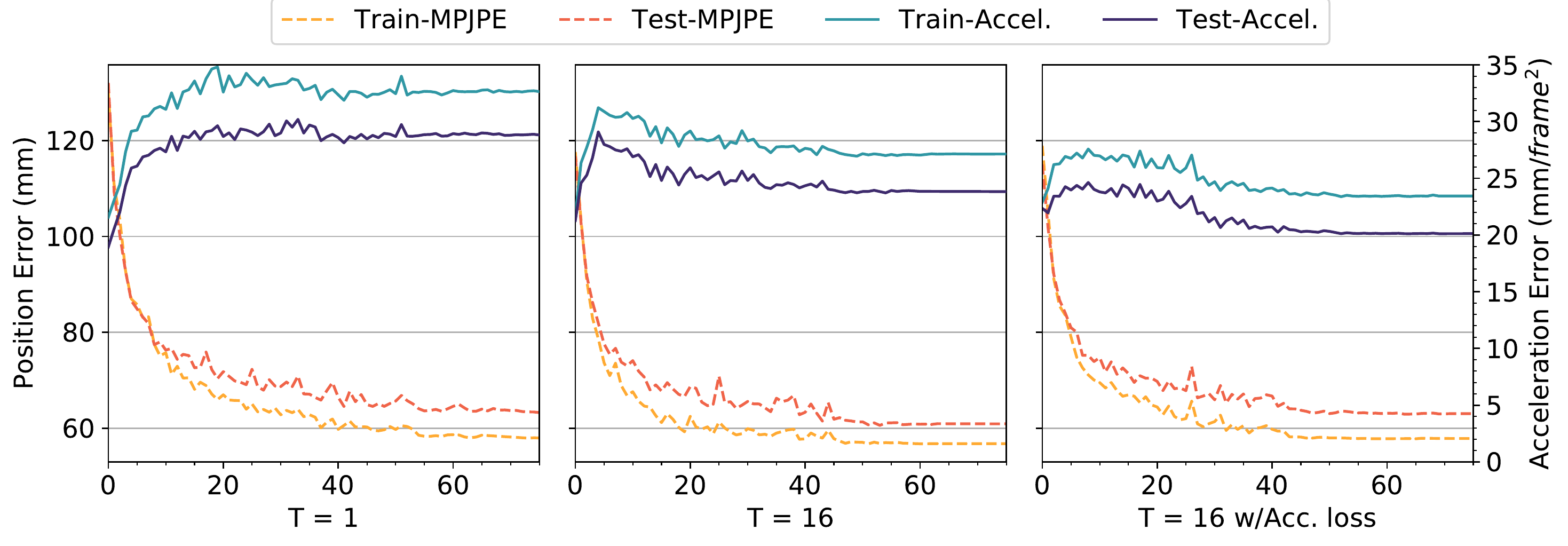}
		\end{minipage}
	}
    	\label{fig:smpl_result}  
\caption{Comparison \textit{MPJPE} and \textit{Accel} during training and testing stages of single-frame ($T = 1$)~\cite{martinez2017simple} and temporal ($T = 27$ or $T = 16$)~\cite{pavllo20193d,kocabas2020vibe} pose estimation and mesh recovery methods. \textit{w/Acc.~loss} adds an acceleration loss in the training stage. In (b) \textit{T = 1}, we simply remove GRUs and set sequence length as $1$.}
\label{fig:motivate} 
\end{figure*}

\subsection{Rethink Existing Spatio-temporal Models}
\label{sec:motivation}
To explore the bottleneck of existing methods using spatio-temporal models to optimize precision and smoothness concurrently, we perform experiments on popular 3D skeleton-based methods~\cite{martinez2017simple,pavllo20193d} and SMPL-based methods~\cite{kolotouros2019spin,kocabas2020vibe}.
In terms of the single-frame approaches ($T = 1$), we implement the simple baseline FCN~\cite{martinez2017simple} for 3D pose estimation tested on the Human3.6M dataset and remove GRUs in VIBE~\cite{kocabas2020vibe} for body recovery tested on the 3DPW dataset. 
For multi-frame methods ($T > 1$), we apply the video-based 3D pose estimator VPose~\cite{pavllo20193d}, conducting temporal convolution networks with dilated convolution along the time axis and the official VIBE~\cite{kocabas2020vibe}. The difference between single-frame methods and multi-frame methods is different from aggregation strategies along the temporal dimension.
Two evaluation metrics, mean position errors (\textit{MPJPE}) and acceleration errors (\textit{Accel}) are used.

Figure~\ref{fig:motivate} illustrates the training and testing performance for both \textit{MPJPE} and \textit{Accel}.
%
For single-frame models ($T = 1$)~\cite{martinez2017simple,kolotouros2019spin}, we observe that the position errors decrease, but the acceleration errors become larger as the epochs increase, indicating that the single-frame methods which extract only spatial information are likely to sacrifice smoothness in exchange for localization performance improvement. It is important to exploit temporal information explicitly.

For multi-frame approaches~\cite{pavllo20193d,kocabas2020vibe} ($T = 27$), they make use of temporal information by TCNs~\cite{pavllo20193d} and GRUs~\cite{kocabas2020vibe} respectively and improve both precision and smoothness.  
Yet, their loss function is applied to each frame, and their smoothness is still far from satisfactory, which is intuitively not beneficial for smoothness optimization.

Accordingly, to further improve smoothness as previous works did~\cite{kanazawa2019learning,Wang2020MotionG3}, we add an acceleration (\textit{Acc.}) loss on the per-frame L1 loss, which constrains the estimated acceleration to be as close as the ground truth's acceleration.
As shown in the right ones ($T = 27$ \textit{w/ Acc.~loss)}, although the acceleration errors decrease slightly, the position errors increase instead. 
It implies that it is hard to achieve optimal precision and smoothness simultaneously within existing frameworks (including models and loss functions).
The reasons behind this may lie in that temporal and spatial information may generalize and overfit at different rates as two different modalities. \textit{MPJPEs} are always larger than \textit{Accels}, making the models pay more attention to optimizing spatial errors and hard to reduce \textit{Accels} greatly.
This observation motivates us to design the \textit{temporal-only refinement paradigm}.

Moreover, to quantitatively explore the combination strategies of \modelname~with existing backbones, whether training two models together (the one-stage strategy) or training them separately (the two-stage method), we try each of them on 3d pose estimation and body recovery. Specifically, if \modelname~is trained together with the backbones in an end-to-end manner (w/ $B$), it belongs to the one-stage strategy. And if \modelname~is trained separately, it is called the two-stage method. As presented in Table~\ref{tab:pose_loss}, we can find that (i) the spatio-temporal model~\cite{pavllo20193d} with multiple frames as inputs will gain in both \textit{Accel} (smoothness) and \textit{MPJPE} (precision), but the computational costs will be increased; (ii) adding acceleration loss or \modelname~in an end-to-end way can benefit \textit{Accel} but harm \textit{MPJPE}; (iii) adding intermediate L1 supervision between the backbones and \modelname~(w/ $B$ $\circ$) shows a slight drop in performance, but after adding an additional acceleration loss will improve both metrics. Compared with one-stage strategies, two-stage solutions with a refinement network show their strengths in boosting both smoothness and precision. 

\begin{table}[h]
	\centering
    \small
    \caption{Comparison results of the body recovery from VIBE~\cite{kocabas2020vibe} of different training strategies on 3DPW. $\times$ means acceleration loss added in the loss function. $B$ means to add \modelname~behind the backbones trained in an end-to-end manner. }
	\scriptsize
	{%
		\begin{tabular}{ll|cccc}
		    
			\specialrule{.1em}{.05em}{.05em}
			&Strategy& \textit{Accel}& \emph{MPJPE}& \emph{PA-MPJPE}& \emph{MPJVE}\\
			\midrule 
			\parbox{2.5mm}{\multirow{5}{*}{\rotatebox[origin=c]{90}{Backbones}}} 
            &In = 1 &32.69&84.54&57.94&102.05\\
            &In = 16 &23.21&83.03&\uline{56.77}&\uline{99.76}\\
            \cmidrule{2-6}
            &In = 16 $\times$ &20.42& 84.51&57.81&101.62\\
            &In = 16 w/ $B$&21.65&86.56&59.93&105.08\\
            \midrule
            \parbox{0.1mm}{\multirow{2}{*}{\rotatebox[origin=c]{90}{Ours}}} 
            &\cellcolor{Gray}In = 1 w/ ours &\cellcolor{Gray}\uline{6.12}&\cellcolor{Gray}\uline{82.98}&\cellcolor{Gray}57.27&\cellcolor{Gray}100.67\\ 
            &\cellcolor{Gray}In = 16 w/ ours &\cellcolor{Gray}\textbf{6.05}&\cellcolor{Gray}\textbf{81.42}&\cellcolor{Gray}\textbf{56.21}&\cellcolor{Gray}\textbf{98.83}\\ 
            
            \midrule
        \end{tabular}%
	}
	\label{tab:vibe_loss}
\end{table}

\begin{table}[h]
	\centering
    \small
    \caption{Comparison of the 3D pose estimation results from VPose~\cite{pavllo20193d} of different training strategies on Human3.6M. $\times$ means acceleration loss added in the loss function. $B$ means to add \modelname~behind the origin network trained in an end-to-end manner. $\circ$ adds an intermediate L1 supervision between the backbone and \modelname.}
	
	\scriptsize
	{%
		\begin{tabular}{ll|cccc}
		    
			\specialrule{.1em}{.05em}{.05em}

			&Strategy& \textit{Accel}& \emph{MPJPE}&  Params.\\
			\midrule 
			\parbox{2.5mm}{\multirow{7}{*}{\rotatebox[origin=c]{90}{Backbones}}} 
            &In = 1 &19.17&54.55&6.39M\\
            &In = 27 & 5.07&50.13& 8.61M\\
            \cmidrule{2-5}
            &In = 27 w/ $\times$ & 4.12&51.48& 8.61M\\
            &In = 27 w/ $B$&2.78&52.65& 8.65M\\
            &In = 27 w/ $B$ $\times$ & 2.87&52.18& 8.65M\\
            &In = 27 w/ $B$ $\circ$ &5.46&51.06& 8.65M\\
            &In = 27 w/ $B$ $\circ$ $\times$ & 2.69&50.94& 8.65M\\
            \midrule
            \parbox{2.5mm}{\multirow{2}{*}{\rotatebox[origin=c]{90}{Ours}}} 
            &\cellcolor{Gray}In = 1 w/ ours &\cellcolor{Gray}\uline{1.03}&\cellcolor{Gray}\uline{52.72}&\cellcolor{Gray}\textbf{0.03M}\\
            &\cellcolor{Gray}In = 27 w/ ours &\cellcolor{Gray}\textbf{0.88}&\cellcolor{Gray}\textbf{50.04}&\cellcolor{Gray}\textbf{0.03M}\\ 
            \midrule
        \end{tabular}%
	}
	\label{tab:pose_loss}
\end{table}

\subsection{More Comparison with Filters}
In main paper Sec. 5.2, we compare the performance with filters on human body recovery.  
We first visualize the qualitative results on a specific axis to demonstrate the effectiveness of \modelname.

\begin{figure}[h]
\begin{center}
\includegraphics[width=0.98\textwidth]{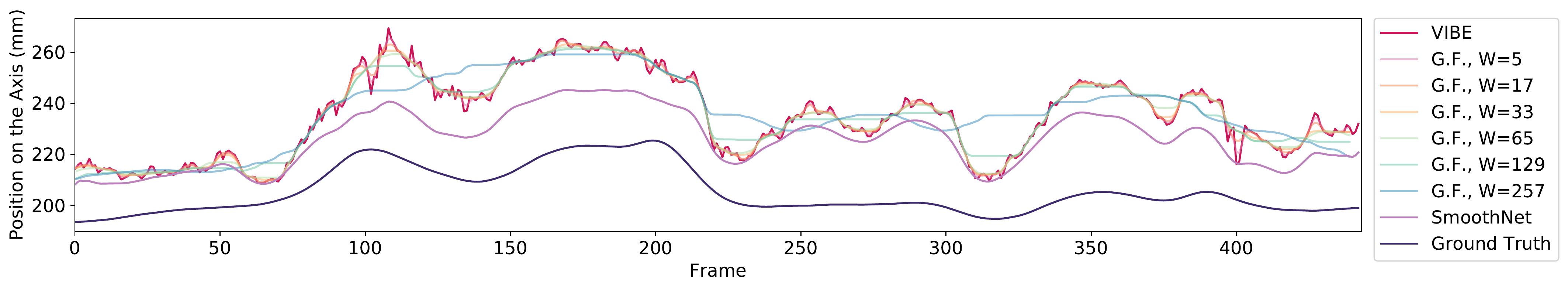}
\end{center}
\caption{Performance comparison between filters and \modelname~on refining the estimated results of VIBE on AIST++ dataset. }
\label{fig:fuse}
\end{figure}

\noindent\textbf{Qualitative comparison.}
Figure~\ref{fig:fuse} illustrates the output positions of VIBE, VIBE with several Gaussian filters (G.F.,) of different kernel sizes, VIBE with our method, and the ground truth. The filters can relieve jitter errors with the increase of kernel size but suffers from over-smoothness when the kernel size is larger than $65$, leading to worse position errors. Instead,
with a learnable design and long-range temporal receptive fields, \modelname~has the capability to learn the long-range noisy patterns and capture more reliable estimations (e.g., near the $70_{th}$ and $220_{th}$ frames) of inputs, making it can not only relieve jitters but also narrow down biased errors consistently.

\noindent\textbf{Quantitative comparison on 2D and 3D pose estimation.}
We further show more results on the tasks of 2D pose estimation and 3D pose estimation on Human3.6M.
In Table~\ref{tab:filter_comp_hm36}, the upper half table of each task compares the results of filters with the closest \emph{MPJPE}s to ours, and the lower half table compares the performance of filters with the most similar \textit{Accel} to ours.
We can conclude that our approach achieves better performance on both precision and smoothness, validating that the temporal-only network with a long-range effective receptive field will be a good solution.

\begin{table}[h]
	\centering
    \small
    \caption{Comparison of most used filters with different estimated poses from CPN~\cite{chen2018cascaded} (2D) and FCN~\cite{martinez2017simple} (3D) on Human3.6M. }
	\scriptsize
	{%
		\begin{tabular}{ll|ccccc}
		    
			\specialrule{.1em}{.05em}{.05em}

			&Method& \textit{Accel} &\emph{MPJPE} &\emph{PA-MPJPE} &Test FPS\\
			
			\midrule

			\parbox{2.5mm}{\multirow{8}{*}{\rotatebox[origin=c]{90}{2D Pose}}}
			 &CPN~\cite{chen2018cascaded}&2.91&6.67&5.18&-\\ 
			 \cmidrule{2-6}
			&w/One-Euro~\cite{casiez20121euro}&0.51&7.86&5.47&2.28k\\
            &w/Savitzky-Golay~\cite{press1990savitzky}&0.20 &6.52&4.99&67.39k\\
            &w/Gaussian1d~\cite{young1995gaus1d}&0.51&6.55&5.00&35.97k\\
            \cmidrule{2-6}
            &w/One-Euro~\cite{casiez20121euro}&0.19&9.21&6.01&3.93k\\
            &w/Savitzky-Golay~\cite{press1990savitzky}&0.15 &8.23&5.89&65.10k\\
            &w/Gaussian1d~\cite{young1995gaus1d}&\textbf{0.14}&6.73&4.99&43.74k\\
            \cmidrule{2-6}
            &\textbf{w/Ours}&\textbf{0.14}&\textbf{ 6.45}&\textbf{ 4.96}&\textbf{71.60k}\\
            		
			\midrule

		    \parbox{2.5mm}{\multirow{8}{*}{\rotatebox[origin=c]{90}{3D Pose}}} 
            &FCN~\cite{martinez2017simple} &19.17&54.55&42.20&-\\
            \cmidrule{2-6}
            &w/One-Euro~\cite{casiez20121euro} &3.80&55.20&42.73&2.27k\\
            &w/Savitzky-Golay~\cite{press1990savitzky}&1.34 &53.48&41.49&66.37k\\
            &w/Gaussian1d~\cite{young1995gaus1d} &2.43&53.67&41.60&29.54k\\
            \cmidrule{2-6}
            &w/One-Euro~\cite{casiez20121euro} &0.94&143.24&85.35&3.72k\\
            &w/Savitzky-Golay~\cite{press1990savitzky}&0.92 &74.38&57.25&65.56k\\
            &w/Gaussian1d~\cite{young1995gaus1d} & 0.95&83.54&68.53&28.93k\\
            \cmidrule{2-6}
            &\textbf{w/Ours}&\textbf{1.03}&\textbf{52.72}&\textbf{40.92}&\textbf{66.67k}\\
        \midrule
        \end{tabular}%
	}
	\label{tab:filter_comp_hm36}
\end{table}


\subsection{Results of Additional Metrics}

To explore the effect on significant errors and long-term jitters, we calculate the worst 1\% of \emph{MPJPE}s (\emph{MPJPE}-1\%) and the worst 1\% \emph{Accel} (\emph{Accel}-1\%) and the corresponding improvement by \modelname. The results of Humane3.6M is shown in the main paper. For the reason that significant errors and long-term jitters are usually accompanied by large estimation errors, as shown in Tab.~ \ref{tab:top1_3D} and Tab.~\ref{tab:top1_smpl}, the improvement on \emph{MPJPE}-1\% and \emph{Accel}-1\% proves the smoothing ability of \modelname on long-term and large jitters.

\renewcommand\arraystretch{1.4}
\begin{table}[H]
\centering
\scriptsize

\caption{\emph{MPJPE}-1\% and \emph{Accel}-1\% improvement on 3D pose estimation. The results of Humane3.6M is shown in the main paper.}

\begin{tabular}{lcccccccc}

\hline
\multicolumn{9}{l}{\textbf{Improvement of \emph{MPJPE}-1\% and \emph{Accel}-1\% on 3D pose estimation}}                                                                                                                                                                                                                                                                                                                                                                                                                                                                                                               \\ \hline
\multicolumn{1}{l|}{Dataset}                                  & \multicolumn{3}{c|}{AIST++}                                                                                                                                                                & \multicolumn{5}{c}{3DPW}                                                                                                                                                                                                                                                                           \\ \hline
\multicolumn{1}{l|}{Estimator}                                & \multicolumn{1}{c|}{SPIN}                                    & \multicolumn{1}{c|}{TCMR}                                    & \multicolumn{1}{c|}{VIBE}                                    & \multicolumn{1}{c|}{EFT}                                     & \multicolumn{1}{c|}{PARE}                                    & \multicolumn{1}{c|}{SPIN}                                    & \multicolumn{1}{c|}{TCMR}                                    & VIBE                                   \\ \hline
\multicolumn{1}{l|}{\emph{MPJPE}-1\%}                                & \multicolumn{1}{c|}{352.93}                                  & \multicolumn{1}{c|}{373.18}                                  & \multicolumn{1}{c|}{339.56}                                  & \multicolumn{1}{c|}{278.51}                                  & \multicolumn{1}{c|}{225.72}                                  & \multicolumn{1}{c|}{289.89}                                  & \multicolumn{1}{c|}{249.11}                                  & 257.83                                 \\
\multicolumn{1}{l|}{\emph{MPJPE}-1\% w/ours} & \multicolumn{1}{c|}{\cellcolor{Gray}\textbf{236.85}} & \multicolumn{1}{c|}{\cellcolor{Gray}\textbf{328.53}} & \multicolumn{1}{c|}{\cellcolor{Gray}\textbf{235.01}} & \multicolumn{1}{c|}{\cellcolor{Gray}\textbf{210.72}} & \multicolumn{1}{c|}{\cellcolor{Gray}\textbf{192.26}} & \multicolumn{1}{c|}{\cellcolor{Gray}\textbf{224.64}} & \multicolumn{1}{c|}{\cellcolor{Gray}\textbf{239.53}} & {\cellcolor{Gray}\textbf{208.26}}                        \\ \hline
\multicolumn{1}{l|}{\emph{Accel}-1\%}                                & \multicolumn{1}{c|}{195.48}                                  & \multicolumn{1}{c|}{43.94}                                   & \multicolumn{1}{c|}{177.07}                                  & \multicolumn{1}{c|}{218.71}                                  & \multicolumn{1}{c|}{132.82}                                  & \multicolumn{1}{c|}{199.33}                                  & \multicolumn{1}{c|}{43.81}                                   & 123.92                                 \\
\multicolumn{1}{l|}{\emph{Accel}-1\% w/ours}                         & \multicolumn{1}{c|}{\cellcolor{Gray}\textbf{12.38}}  & \multicolumn{1}{c|}{\cellcolor{Gray}\textbf{12.30}}  & \multicolumn{1}{c|}{\cellcolor{Gray}\textbf{11.98}}  & \multicolumn{1}{c|}{\cellcolor{Gray}\textbf{29.73}}  & \multicolumn{1}{c|}{\cellcolor{Gray}\textbf{31.75}}  & \multicolumn{1}{c|}{\cellcolor{Gray}\textbf{25.88}}  & \multicolumn{1}{c|}{\cellcolor{Gray}\textbf{30.80}}  & {\cellcolor{Gray}\textbf{26.11}} \\ \hline\hline
\multicolumn{1}{l|}{Dataset}                                  & \multicolumn{3}{c|}{MPI-INF-3DHP}                                                                                                                                                          & \multicolumn{5}{c}{MuPoTS}                                                                                                                                                                                                                                                                         \\ \hline
\multicolumn{1}{l|}{Estimator}                                & \multicolumn{1}{c|}{SPIN}                                    & \multicolumn{1}{c|}{TCMR}                                    & \multicolumn{1}{c|}{VIBE}                                    & \multicolumn{2}{c|}{TposeNet}                                                                                               & \multicolumn{3}{c}{TposeNet w/RefineNet}                                                                                                                             \\ \hline
\multicolumn{1}{l|}{\emph{MPJPE}-1\%}                                & \multicolumn{1}{c|}{273.73}                                  & \multicolumn{1}{c|}{255.93}                                  & \multicolumn{1}{c|}{253.58}                                  & \multicolumn{2}{c|}{354.87}                                                                                                 & \multicolumn{3}{c}{277.79}                                                                                                                                           \\
\multicolumn{1}{l|}{\emph{MPJPE}-1\% w/ours}                         & \multicolumn{1}{c|}{\cellcolor{Gray}\textbf{241.62}} & \multicolumn{1}{c|}{\cellcolor{Gray}\textbf{246.64}} & \multicolumn{1}{c|}{\cellcolor{Gray}\textbf{238.57}} & \multicolumn{2}{c|}{\cellcolor{Gray}\textbf{347.36}}                                                                & \multicolumn{3}{c}{\cellcolor{Gray}\textbf{265.98}}                                                                                                          \\ \hline
\multicolumn{1}{l|}{\emph{Accel}-1\%}                                & \multicolumn{1}{c|}{107.00}                                  & \multicolumn{1}{c|}{16.73}                                   & \multicolumn{1}{c|}{57.05}                                   & \multicolumn{2}{c|}{33.75}                                                                                                  & \multicolumn{3}{c}{26.65}                                                                                                                                            \\
\multicolumn{1}{l|}{\emph{Accel}-1\% w/ours}                         & \multicolumn{1}{c|}{\cellcolor{Gray}\textbf{9.89}}   & \multicolumn{1}{c|}{\cellcolor{Gray}\textbf{9.24}}   & \multicolumn{1}{c|}{\cellcolor{Gray}\textbf{9.11}}   & \multicolumn{2}{c|}{\cellcolor{Gray}\textbf{6.17}}                                                                  & \multicolumn{3}{c}{\cellcolor{Gray}\textbf{8.75}}                                                                                                            \\ \hline
\end{tabular}
	\label{tab:top1_3D}
\end{table}

\begin{table}[]
\centering
\scriptsize

\caption{\emph{MPJPE}-1\% and \emph{Accel}-1\% improvement on SMPL pose estimation.}
\begin{tabular}{lcccccccc}
\hline
\multicolumn{9}{l}{\textbf{Improvement of \emph{MPJPE}-1\% and \emph{Accel}-1\% on SMPL Pose}}                                                                                                                                                                                                                                                       \\ \hline
\multicolumn{1}{l|}{Dataset}                                  & \multicolumn{3}{c|}{AIST++}                                                                                                                                                                & \multicolumn{5}{c}{3DPW}                                                                                                                                                                                                                                                                           \\ \hline
\multicolumn{1}{l|}{Estimator}                                & \multicolumn{1}{c|}{SPIN}                                    & \multicolumn{1}{c|}{TCMR}                                    & \multicolumn{1}{c|}{VIBE}                                    & \multicolumn{1}{c|}{EFT}                                     & \multicolumn{1}{c|}{PARE}                                    & \multicolumn{1}{c|}{SPIN}                                    & \multicolumn{1}{c|}{TCMR}                                    & VIBE                                   \\ \hline
\multicolumn{1}{l|}{\emph{MPJPE}-1\%}                                & \multicolumn{1}{c|}{355.85}                                  & \multicolumn{1}{c|}{374.36}                                  & \multicolumn{1}{c|}{341.80}                                  & \multicolumn{1}{c|}{272.32}                                  & \multicolumn{1}{c|}{232.71}                                  & \multicolumn{1}{c|}{283.56}                                  & \multicolumn{1}{c|}{251.94}                                  & 255.49                                 \\

\multicolumn{1}{l|}{\emph{MPJPE}-1\% w/ours} & \multicolumn{1}{c|}{\cellcolor{Gray}\textbf{270.94}} & \multicolumn{1}{c|}{\cellcolor{Gray}\textbf{352.54}} & \multicolumn{1}{c|}{\cellcolor{Gray}\textbf{274.19}} & \multicolumn{1}{c|}{\cellcolor{Gray}\textbf{223.91}} & \multicolumn{1}{c|}{\cellcolor{Gray}\textbf{207.05}} & \multicolumn{1}{c|}{\cellcolor{Gray}\textbf{236.66}} & \multicolumn{1}{c|}{\cellcolor{Gray}\textbf{249.34}} & {\cellcolor{Gray}\textbf{218.57}}                        \\ \hline
\multicolumn{1}{l|}{\emph{Accel}-1\%}                                & \multicolumn{1}{c|}{195.00}                                  & \multicolumn{1}{c|}{44.13}                                   & \multicolumn{1}{c|}{176.63}                                  & \multicolumn{1}{c|}{205.32}                                  & \multicolumn{1}{c|}{130.74}                                  & \multicolumn{1}{c|}{185.50}                                  & \multicolumn{1}{c|}{38.26}                                   & 118.80                                 \\
\multicolumn{1}{l|}{\emph{Accel}-1\% w/ours}                         & \multicolumn{1}{c|}{\cellcolor{Gray}\textbf{19.97}}  & \multicolumn{1}{c|}{\cellcolor{Gray}\textbf{24.51}}  & \multicolumn{1}{c|}{\cellcolor{Gray}\textbf{23.19}}  & \multicolumn{1}{c|}{\cellcolor{Gray}\textbf{42.32}}  & \multicolumn{1}{c|}{\cellcolor{Gray}\textbf{31.96}}  & \multicolumn{1}{c|}{\cellcolor{Gray}\textbf{33.18}}  & \multicolumn{1}{c|}{\cellcolor{Gray}\textbf{36.39}}  & \cellcolor{Gray}\textbf{31.28} \\ \hline
\end{tabular}

	\label{tab:top1_smpl}

\end{table}

\renewcommand\arraystretch{1}

\subsection{Comparison with RefineNet}
\label{sec:multi}

For learning-based jitter mitigation methods, we choose RefineNet~\cite{veges2020temporal} for comparison on the multi-person 3D pose estimation dataset MuPoTS-3D~\cite{Mehta2018SingleShotM3}. We compare them on the universal coordinates, where each person is rescaled according to the hip and has a normalized height. We also show the refinement results with two filters for comparison. 
As RefineNet~\cite{veges2020temporal} has compared with the interpolation methods and One-Euro filter~\cite{casiez20121euro} and showed better performance, we do not list their results here.
To better fit the test set MuPoTS-3D, RefineNet is trained on two multi-person 3D poses datasets: MPI-INF-3DHP dataset~\cite{mehta2017monocular} and an in-distribution MuCo-Temp dataset generated by the authors. 
%
%
In contrast, \modelname~is trained on VIBE-AIST++ (the same model used in the previous experiment) without any finetuning to explore the generalization capability of \modelname~across datasets.

\begin{table}[h]
	\centering
    \small
    \caption{\textbf{Comparison results on multi-person MuPoTS-3D dataset~\cite{Mehta2018SingleShotM3}.} \modelname~is directly tested on it, while RefineNet~\cite{veges2020temporal} has been trained on in-domain datasets.}
 
	{%
		\begin{tabular}{l|ccc}

			\specialrule{.1em}{.05em}{.05em}
			Method& \emph{Accel} & \emph{MPJPE}& \emph{PA-MPJPE}\\
			\midrule
			TPoseNet~\cite{veges2020temporal}& 12.70&103.33&68.36\\
			TPoseNet w/ RefineNet~\cite{veges2020temporal} &9.53&\textbf{93.97}&\textbf{65.16}\\
			TPoseNet w/ Savitzky-Golay &\underline{8.29}&102.79&68.30\\
			TPoseNet w/ Gaussian1d&8.61&102.70&68.17\\
			\cellcolor{Gray}TPoseNet w/ Ours &\cellcolor{Gray}\textbf{7.23}&\cellcolor{Gray}\underline{100.78}&\cellcolor{Gray}\underline{68.10}\\
			\midrule
            RefineNet w/ Savitzky-Golay &\underline{7.22}&93.75&65.34\\
            RefineNet w/ Gaussian1d &8.40&\underline{93.65}&\underline{65.19}\\
            \cellcolor{Gray}RefineNet w/ Ours &\cellcolor{Gray}\textbf{7.21}&\cellcolor{Gray}\textbf{91.78}&\cellcolor{Gray}\textbf{65.06}\\

        \midrule
        \end{tabular}%
	}
	\label{tab:refine}
\end{table}

In Table~\ref{tab:refine}, we first analyze the refinement results for the TPoseNet pose estimator~\cite{veges2020temporal}, which is a temporal residual convolutional network for 2D-to-3D pose estimation used as the backbone network in RefineNet. 
Although the \emph{MPJPE} of RefineNet drops the most as it has been trained on the relevant datasets, its \emph{Accel} is the highest, indicating that the smoothing capability of RefineNet cannot outperform filter-based solutions~\cite{young1995gaus1d,press1990savitzky}. 
Our method further improves on \emph{Accel} by $8.35$\% compared to the best filter solution. At the same time, as a data-driven method, even though \modelname~is trained on a different dataset, it shows a $1.9$\% reduction in pose estimation errors compared to the motion-oblivious filter-based solutions.

Finally, it is possible to refine the pose outputs from RefineNet with filters and~\modelname, and we show the results in the bottom half of Table~\ref{tab:refine}.
As observed from the table, all the methods result in performance improvement. Among them, \modelname~again obtains the largest improvements, \textit{i.e.}, $24.3$\% and $2.3$\% in \emph{Accel} and \emph{MPJPE}, respectively. Such results demonstrate the effectiveness of the proposed solution on top of any learning-based pose estimators.

\subsection{Generalization Ability}

\modelname is a temporal-only network, which has good generalization ability across backbones, modalities, and datasets. By default, we provide three pretrained models (Trained 3D positions on the FCN estimator on Human3.6M, SPIN on 3DPW \cite{von2018recovering}, and VIBE on AIST++ \cite{li2021aist}), covering all existing cases and making \emph{Accel}s reduce greatly. Due to motion distribution shift, the results of \emph{MPJPEs} and \emph{PA-MPJPEs} will be influenced. 

To explore the generalization ability, we present test results (shown in the left two columns) under the three pretrained models (shown in the third row) in Table~\ref{tab:supp_gd_smpl}, ~\ref{tab:supp_gd_2d}, ~\ref{tab:supp_gd_3d}. The background color highlights the increase and decrease degree \modelname outputs have compared with inputs. The greener the background is, the better output result is. The redder the background is, the worse the output reulst is. We have some observations as follows. First, for all models, the \emph{Accels} are lower than INPUT estimators by a large margin (illustrated as green background), indicating that they have similar smoothing capability. Second, for \emph{MPJPEs} and \emph{PA-MPJPEs}, we can find at least one model with better accuracy than results from INPUT estimators. Third, the \emph{MPJPEs} will be lower if the trained datasets are the same or have similar distribution to the test datasets.

\begin{table}[H]
\tiny
\caption{\textbf{Comparison results on generalization ability across modality (from 3d position to 6d rotation matrix)} and backbones (shown in the left two columns) under the three pretrained \modelname (shown in the third row). Green background means great improvement, while red background highlights the worse results compared with Input estimators.}
\begin{tabular}{llcccccccccccc}
\hline
\multicolumn{14}{c}{\textbf{Generalization Ability on SMPL Pose Estimation}}                                                                                                                                        \\ \hline
\multicolumn{2}{c|}{Mertrics}                                                                                                                                            & \multicolumn{4}{c|}{Accel}                                                                                                                            & \multicolumn{4}{c|}{MPJPE}                                                                                                                                  & \multicolumn{4}{c}{PA-MPJPE}                                                                                                                                      \\ \hline
\multicolumn{1}{l|}{\begin{tabular}[c]{@{}l@{}}Test\\ Dataset\end{tabular}} & \multicolumn{1}{l|}{\begin{tabular}[c]{@{}l@{}}Train\\ \textbackslash\\ Test\end{tabular}} & \multicolumn{1}{c|}{INPUT}    & \multicolumn{1}{c|}{AIST++}           & \multicolumn{1}{c|}{H36M}             & \multicolumn{1}{c|}{3DPW}             & \multicolumn{1}{c|}{INPUT}     & \multicolumn{1}{c|}{AIST++}             & \multicolumn{1}{c|}{H36M}               & \multicolumn{1}{c|}{3DPW}              & \multicolumn{1}{c|}{INPUT}             & \multicolumn{1}{c|}{AIST++}            & \multicolumn{1}{c|}{H36M}              & 3DPW                                   \\ \hline
\multicolumn{1}{l|}{}                                                       & \multicolumn{1}{l|}{SPIN}                                                                  & \cellcolor[HTML]{FFF2CC}33.21 & \cellcolor[HTML]{57BB8A}\textbf{5.72} & \cellcolor[HTML]{67C195}5.81          & \cellcolor[HTML]{B5E1CC}6.24          & \cellcolor[HTML]{FFF2CC}107.72 & \cellcolor[HTML]{57BB8A}\textbf{103.00} & \cellcolor[HTML]{ABDDC5}104.03          & \cellcolor[HTML]{FFFFFF}105.04         & \cellcolor[HTML]{FFF2CC}74.39          & \cellcolor[HTML]{57BB8A}\textbf{70.98} & \cellcolor[HTML]{C5E7D6}71.69          & \cellcolor[HTML]{FFFFFF}72.06          \\ \cline{2-14} 
\multicolumn{1}{l|}{}                                                       & \multicolumn{1}{l|}{{\color[HTML]{333333} TCMR}}                                           & \cellcolor[HTML]{FFF2CC}6.47  & \cellcolor[HTML]{78C8A1}4.70          & \cellcolor[HTML]{57BB8A}\textbf{4.68} & \cellcolor[HTML]{78C8A1}4.70          & \cellcolor[HTML]{FFF2CC}106.95 & \cellcolor[HTML]{E67C73}108.19          & \cellcolor[HTML]{57BB8A}\textbf{106.39} & \cellcolor[HTML]{FBE6E4}107.19         & \cellcolor[HTML]{FFF2CC}71.58          & \cellcolor[HTML]{E67C73}71.89          & \cellcolor[HTML]{57BB8A}\textbf{71.33} & \cellcolor[HTML]{9AD6B8}71.43          \\ \cline{2-14} 
\multicolumn{1}{l|}{\multirow{-3}{*}{AIST++}}                               & \multicolumn{1}{l|}{VIBE}                                                                  & \cellcolor[HTML]{FFF2CC}31.65 & \cellcolor[HTML]{57BB8A}\textbf{5.88} & \cellcolor[HTML]{65C093}5.95          & \cellcolor[HTML]{B5E1CB}6.34          & \cellcolor[HTML]{FFF2CC}107.41 & \cellcolor[HTML]{57BB8A}\textbf{102.06} & \cellcolor[HTML]{D8EFE3}104.48          & \cellcolor[HTML]{FFFFFF}105.21         & \cellcolor[HTML]{FFF2CC}72.83          & \cellcolor[HTML]{FFFFFF}69.49          & \cellcolor[HTML]{57BB8A}\textbf{70.38} & \cellcolor[HTML]{FFFFFF}70.74          \\ \hline
\multicolumn{1}{l|}{}                                                       & \multicolumn{1}{l|}{EFT}                                                                   & \cellcolor[HTML]{FFF2CC}33.38 & \cellcolor[HTML]{57BB8A}\textbf{7.67} & \cellcolor[HTML]{67C195}7.71          & \cellcolor[HTML]{AEDEC7}7.89          & \cellcolor[HTML]{FFF2CC}91.6   & \cellcolor[HTML]{E67C73}91.79           & \cellcolor[HTML]{A7DBC1}90.54           & \cellcolor[HTML]{57BB8A}\textbf{89.57} & \cellcolor[HTML]{FFF2CC}55.33          & \cellcolor[HTML]{C5E7D7}54.96          & \cellcolor[HTML]{FFFFFF}\textbf{55.25} & \cellcolor[HTML]{57BB8A}\textbf{54.40} \\ \cline{2-14} 
\multicolumn{1}{l|}{}                                                       & \multicolumn{1}{l|}{PARE}                                                                  & \cellcolor[HTML]{FFF2CC}26.45 & \cellcolor[HTML]{60BE90}6.29          & \cellcolor[HTML]{57BB8A}\textbf{6.28} & \cellcolor[HTML]{72C69D}6.31          & \cellcolor[HTML]{FFF2CC}79.93  & \cellcolor[HTML]{E67C73}81.37           & \cellcolor[HTML]{FFFFFF}79.93           & \cellcolor[HTML]{57BB8A}\textbf{78.68} & \cellcolor[HTML]{FFF2CC}48.74          & \cellcolor[HTML]{F2BAB5}49.14          & \cellcolor[HTML]{E67C73}49.49          & \cellcolor[HTML]{57BB8A}\textbf{48.47} \\ \cline{2-14} 
\multicolumn{1}{l|}{}                                                       & \multicolumn{1}{l|}{SPIN}                                                                  & \cellcolor[HTML]{FFF2CC}34.95 & \cellcolor[HTML]{57BB8A}\textbf{7.22} & \cellcolor[HTML]{57BB8A}\textbf{7.22} & \cellcolor[HTML]{A6DBC1}7.40          & \cellcolor[HTML]{FFF2CC}99.28  & \cellcolor[HTML]{FFFFFF}98.98           & \cellcolor[HTML]{83CDA9}98.12           & \cellcolor[HTML]{57BB8A}\textbf{97.81} & \cellcolor[HTML]{FFF2CC}61.71          & \cellcolor[HTML]{F8D7D4}61.79          & \cellcolor[HTML]{E67C73}61.97          & \cellcolor[HTML]{57BB8A}\textbf{61.19} \\ \cline{2-14} 
\multicolumn{1}{l|}{}                                                       & \multicolumn{1}{l|}{TCMR}                                                                  & \cellcolor[HTML]{FFF2CC}7.12  & \cellcolor[HTML]{8ED1B0}6.50          & \cellcolor[HTML]{C6E8D7}6.52          & \cellcolor[HTML]{57BB8A}\textbf{6.48} & \cellcolor[HTML]{FFF2CC}88.46  & \cellcolor[HTML]{E67C73}89.82           & \cellcolor[HTML]{57BB8A}\textbf{88.37}  & \cellcolor[HTML]{FBE9E8}88.69          & \cellcolor[HTML]{FFF2CC}\textbf{55.70} & \cellcolor[HTML]{E67C73}57.22          & \cellcolor[HTML]{FFFFFF}55.97          & \cellcolor[HTML]{F3BCB8}56.61          \\ \cline{2-14} 
\multicolumn{1}{l|}{\multirow{-5}{*}{3DPW}}                                 & \multicolumn{1}{l|}{VIBE}                                                                  & \cellcolor[HTML]{FFF2CC}23.59 & \cellcolor[HTML]{57BB8A}\textbf{5.98} & \cellcolor[HTML]{E1F3EA}7.24          & \cellcolor[HTML]{F5FBF8}7.42          & \cellcolor[HTML]{FFF2CC}84.27  & \cellcolor[HTML]{E67C73}85.89           & \cellcolor[HTML]{57BB8A}\textbf{83.14}  & \cellcolor[HTML]{86CEAB}83.46          & \cellcolor[HTML]{FFF2CC}54.92          & \cellcolor[HTML]{57BB8A}\textbf{52.49} & \cellcolor[HTML]{EEF8F3}54.6           & \cellcolor[HTML]{FFFFFF}54.83          \\ \hline
\end{tabular}
\label{tab:supp_gd_smpl}
\end{table}

\renewcommand\arraystretch{1.4}

\begin{table}[H]
\tiny
\caption{\textbf{Comparison results on generalization ability across modality (from 3d position to 2d position)} and backbones (shown in the left two columns) under the three pretrained \modelname (shown in the third row). }
\begin{tabular}{llcccccccccccc}
\hline
\multicolumn{14}{c}{\textbf{Generalization Ability on 2D Pose Estimation}}                                                                      \\ \hline
\multicolumn{2}{l|}{Mertrics}                                                                                                                                            & \multicolumn{4}{c|}{Accel}                                                                                                                                                & \multicolumn{4}{c|}{MPJPE}                                                                                                          & \multicolumn{4}{c}{PA-MPJPE}                                                                                                                \\ \hline
\multicolumn{1}{l|}{\begin{tabular}[c]{@{}l@{}}Test\\ Dataset\end{tabular}} & \multicolumn{1}{l|}{\begin{tabular}[c]{@{}l@{}}Train\\ \textbackslash\\ Test\end{tabular}} & \multicolumn{1}{c|}{INPUT}   & \multicolumn{1}{c|}{AIST++}           & \multicolumn{1}{c|}{H36M}             & \multicolumn{1}{c|}{3DPW}                                  & \multicolumn{1}{c|}{INPUT}   & \multicolumn{1}{c|}{AIST++}   & \multicolumn{1}{c|}{H36M}             & \multicolumn{1}{c|}{3DPW}    & \multicolumn{1}{c|}{INPUT}   & \multicolumn{1}{c|}{AIST++}  & \multicolumn{1}{c|}{H36M}             & 3DPW                                  \\ \hline
\multicolumn{1}{l|}{}                                                       & \multicolumn{1}{l|}{\cellcolor[HTML]{FFFFFF}CPN}                                           & \cellcolor[HTML]{FFF2CC}2.91 & \cellcolor[HTML]{7CCAA3}0.16          & \cellcolor[HTML]{57BB8A}\textbf{0.14} & \multicolumn{1}{c|}{\cellcolor[HTML]{8FD1B1}0.17}          & \cellcolor[HTML]{FFF2CC}6.67 & \cellcolor[HTML]{E67C73}8.52  & \cellcolor[HTML]{57BB8A}\textbf{6.45} & \cellcolor[HTML]{EC9B94}8.09 & \cellcolor[HTML]{FFF2CC}5.18 & \cellcolor[HTML]{E67C73}5.88 & \cellcolor[HTML]{57BB8A}\textbf{4.96} & \cellcolor[HTML]{F6CFCB}5.44          \\ \cline{2-14} 
\multicolumn{1}{l|}{}                                                       & \multicolumn{1}{l|}{\cellcolor[HTML]{FFFFFF}Hourglass}                                     & \cellcolor[HTML]{FFF2CC}1.54 & \cellcolor[HTML]{57BB8A}\textbf{0.15} & \cellcolor[HTML]{57BB8A}\textbf{0.15} & \multicolumn{1}{c|}{\cellcolor[HTML]{81CCA7}0.16}          & \cellcolor[HTML]{FFF2CC}9.42 & \cellcolor[HTML]{E67C73}10.64 & \cellcolor[HTML]{57BB8A}\textbf{9.25} & \cellcolor[HTML]{F5CAC6}9.92 & \cellcolor[HTML]{FFF2CC}7.64 & \cellcolor[HTML]{E67C73}8.3  & \cellcolor[HTML]{B0DFC8}7.57          & \cellcolor[HTML]{57BB8A}\textbf{7.49} \\ \cline{2-14} 
\multicolumn{1}{l|}{}                                                       & \multicolumn{1}{l|}{\cellcolor[HTML]{FFFFFF}HRNet}                                         & \cellcolor[HTML]{FFF2CC}1.01 & \cellcolor[HTML]{57BB8A}\textbf{0.13} & \cellcolor[HTML]{57BB8A}\textbf{0.13} & \multicolumn{1}{c|}{\cellcolor[HTML]{ABDDC4}0.14}          & \cellcolor[HTML]{FFF2CC}4.59 & \cellcolor[HTML]{E67C73}7.09  & \cellcolor[HTML]{57BB8A}\textbf{4.54} & \cellcolor[HTML]{EDA19B}6.39 & \cellcolor[HTML]{FFF2CC}4.19 & \cellcolor[HTML]{E67C73}5.47 & \cellcolor[HTML]{57BB8A}\textbf{4.13} & \cellcolor[HTML]{F7D2CF}4.63          \\ \cline{2-14} 
\multicolumn{1}{l|}{\multirow{-4}{*}{H36M}}                            & \multicolumn{1}{l|}{\cellcolor[HTML]{FFFFFF}RLE}                                           & \cellcolor[HTML]{FFF2CC}0.9  & \cellcolor[HTML]{57BB8A}\textbf{0.13} & \cellcolor[HTML]{57BB8A}\textbf{0.13} & \multicolumn{1}{c|}{\cellcolor[HTML]{57BB8A}\textbf{0.13}} & \cellcolor[HTML]{FFF2CC}5.14 & \cellcolor[HTML]{E67C73}7.67  & \cellcolor[HTML]{57BB8A}\textbf{5.11} & \cellcolor[HTML]{EB968F}7.18 & \cellcolor[HTML]{FFF2CC}4.82 & \cellcolor[HTML]{E67C73}6.04 & \cellcolor[HTML]{57BB8A}\textbf{4.78} & \cellcolor[HTML]{F6CCC8}5.3           \\ \hline
\end{tabular}
\label{tab:supp_gd_2d}
\end{table}

\begin{table}[H]
\tiny
\caption{\textbf{Comparison results on generalization ability across backbone} under the three pretrained \modelname (shown in the third row). }
\begin{tabular}{llcccccccccccc}
\hline
\multicolumn{14}{c}{\textbf{Generalization Ability on 3D Pose Estimation}}                                                                                                                                                                                      \\ \hline
\multicolumn{2}{c|}{Evaluation Mertrics}                                                                                                                                                    & \multicolumn{4}{c|}{Accel}                                                                                                                            & \multicolumn{4}{c|}{MPJPE}                                                                                                                                          & \multicolumn{4}{c}{PA-MPJPE}                                                                                                                                      \\ \hline
\multicolumn{1}{l|}{\begin{tabular}[c]{@{}l@{}}Test\\ Dataset\end{tabular}}                    & \multicolumn{1}{l|}{\begin{tabular}[c]{@{}l@{}}Train\\ \textbackslash\\ Test\end{tabular}} & \multicolumn{1}{c|}{INPUT}    & \multicolumn{1}{c|}{AIST}             & \multicolumn{1}{c|}{H36M}             & \multicolumn{1}{c|}{3DPW}             & \multicolumn{1}{c|}{INPUT}             & \multicolumn{1}{c|}{AIST}               & \multicolumn{1}{c|}{H36M}              & \multicolumn{1}{c|}{3DPW}               & \multicolumn{1}{c|}{INPUT}             & \multicolumn{1}{c|}{AIST}              & \multicolumn{1}{c|}{H36M}              & 3DPW                                   \\ \hline
\multicolumn{1}{l|}{}                                                                          & \multicolumn{1}{l|}{SPIN}                                                                  & \cellcolor[HTML]{FFF2CC}33.19 & \cellcolor[HTML]{57BB8A}\textbf{4.17} & \cellcolor[HTML]{64C093}4.23          & \cellcolor[HTML]{ABDDC4}4.54          & \cellcolor[HTML]{FFF2CC}107.17         & \cellcolor[HTML]{57BB8A}\textbf{95.21}  & \cellcolor[HTML]{FFFFFF}104.03         & \cellcolor[HTML]{C3E6D5}100.91          & \cellcolor[HTML]{FFF2CC}74.40          & \cellcolor[HTML]{57BB8A}\textbf{69.98} & \cellcolor[HTML]{FFFFFF}72.71          & \cellcolor[HTML]{DFF2E9}72.20          \\ \cline{2-14} 
\multicolumn{1}{l|}{}                                                                          & \multicolumn{1}{l|}{{\color[HTML]{333333} TCMR}}                                           & \cellcolor[HTML]{FFF2CC}6.40  & \cellcolor[HTML]{81CCA7}4.24          & \cellcolor[HTML]{73C69D}4.23          & \cellcolor[HTML]{57BB8A}\textbf{4.21} & \cellcolor[HTML]{FFF2CC}106.72         & \cellcolor[HTML]{57BB8A}\textbf{105.51} & \cellcolor[HTML]{CAE9DA}105.84         & \cellcolor[HTML]{FFFFFF}105.99          & \cellcolor[HTML]{FFF2CC}\textbf{71.59} & \cellcolor[HTML]{E67C73}72.23          & \cellcolor[HTML]{FFFFFF}71.61          & \cellcolor[HTML]{F4C2BE}71.90          \\ \cline{2-14} 
\multicolumn{1}{l|}{\multirow{-3}{*}{AIST}}                                                    & \multicolumn{1}{l|}{VIBE}                                                                  & \cellcolor[HTML]{FFF2CC}31.64 & \cellcolor[HTML]{57BB8A}\textbf{4.15} & \cellcolor[HTML]{68C196}4.22          & \cellcolor[HTML]{ADDDC6}4.50          & \cellcolor[HTML]{FFF2CC}106.90         & \cellcolor[HTML]{57BB8A}\textbf{97.47}  & \cellcolor[HTML]{FFFFFF}104.01         & \cellcolor[HTML]{EDF7F2}103.31          & \cellcolor[HTML]{FFF2CC}72.84          & \cellcolor[HTML]{FFFFFF}69.67          & \cellcolor[HTML]{57BB8A}\textbf{71.26} & \cellcolor[HTML]{FFFFFF}70.89          \\ \hline
\multicolumn{1}{l|}{}                                                                          & \multicolumn{1}{l|}{FCN}                                                                   & \cellcolor[HTML]{FFF2CC}19.17 & \cellcolor[HTML]{67C195}1.07          & \cellcolor[HTML]{57BB8A}\textbf{1.03} & \cellcolor[HTML]{9CD7BA}1.20          & \cellcolor[HTML]{FFF2CC}54.55          & \cellcolor[HTML]{E67C73}61.79           & \cellcolor[HTML]{57BB8A}\textbf{52.72} & \cellcolor[HTML]{F3BDB9}\textbf{58.22}  & \cellcolor[HTML]{FFF2CC}42.20          & \cellcolor[HTML]{E67C73}43.85          & \cellcolor[HTML]{57BB8A}\textbf{40.92} & \cellcolor[HTML]{D3EDE0}41.87          \\ \cline{2-14} 
\multicolumn{1}{l|}{}                                                                          & \multicolumn{1}{l|}{RLE}                                                                   & \cellcolor[HTML]{FFF2CC}7.75  & \cellcolor[HTML]{62BF91}0.91          & \cellcolor[HTML]{57BB8A}\textbf{0.90} & \cellcolor[HTML]{8ED1B0}0.95          & \cellcolor[HTML]{FFF2CC}48.87          & \cellcolor[HTML]{E67C73}53.29           & \cellcolor[HTML]{57BB8A}\textbf{48.27} & \cellcolor[HTML]{F2B8B3}\textbf{51.28}  & \cellcolor[HTML]{FFF2CC}38.63          & \cellcolor[HTML]{E67C73}40.54          & \cellcolor[HTML]{57BB8A}\textbf{38.13} & \cellcolor[HTML]{BBE3D0}38.43          \\ \cline{2-14} 
\multicolumn{1}{l|}{}                                                                          & \multicolumn{1}{l|}{TCMR}                                                                  & \cellcolor[HTML]{FFF2CC}3.77  & \cellcolor[HTML]{C6E8D7}2.80          & \cellcolor[HTML]{8FD1B1}\textbf{2.79} & \cellcolor[HTML]{FFFFFF}2.81          & \cellcolor[HTML]{FFF2CC}\textbf{73.57} & \cellcolor[HTML]{E67C73}80.43           & \cellcolor[HTML]{FFFFFF}73.89          & \cellcolor[HTML]{E8857D}\textbf{79.99}  & \cellcolor[HTML]{FFF2CC}\textbf{52.04} & \cellcolor[HTML]{E67C73}53.72          & \cellcolor[HTML]{FFFFFF}52.13          & \cellcolor[HTML]{ED9D96}\textbf{53.33} \\ \cline{2-14} 
\multicolumn{1}{l|}{}                                                                          & \multicolumn{1}{l|}{VIBE}                                                                  & \cellcolor[HTML]{FFF2CC}15.81 & \cellcolor[HTML]{57BB8A}\textbf{2.86} & \cellcolor[HTML]{57BB8A}\textbf{2.86} & \cellcolor[HTML]{B8E2CD}2.97          & \cellcolor[HTML]{FFF2CC}78.10          & \cellcolor[HTML]{E67C73}84.89           & \cellcolor[HTML]{57BB8A}\textbf{77.23} & \cellcolor[HTML]{EEA29C}82.94           & \cellcolor[HTML]{FFF2CC}53.67          & \cellcolor[HTML]{E67C73}54.85          & \cellcolor[HTML]{57BB8A}\textbf{53.35} & \cellcolor[HTML]{F7D4D1}54.06          \\ \cline{2-14} 
\multicolumn{1}{l|}{}                                                                          & \multicolumn{1}{l|}{\begin{tabular}[c]{@{}l@{}}VideoPose\\ (T=243)\end{tabular}}           & \cellcolor[HTML]{FFF2CC}2.82  & \cellcolor[HTML]{81CCA7}0.88          & \cellcolor[HTML]{57BB8A}\textbf{0.87} & \cellcolor[HTML]{81CCA7}0.88          & \cellcolor[HTML]{FFF2CC}48.11          & \cellcolor[HTML]{E67C73}53.93           & \cellcolor[HTML]{57BB8A}\textbf{48.05} & \cellcolor[HTML]{EC9B94}52.58           & \cellcolor[HTML]{FFF2CC}37.71          & \cellcolor[HTML]{E67C73}39.74          & \cellcolor[HTML]{57BB8A}\textbf{37.66} & \cellcolor[HTML]{FAE2E0}38.16          \\ \cline{2-14} 
\multicolumn{1}{l|}{}                                                                          & \multicolumn{1}{l|}{\begin{tabular}[c]{@{}l@{}}VideoPose\\ (T=27)\end{tabular}}            & \cellcolor[HTML]{FFF2CC}3.53  & \cellcolor[HTML]{87CEAB}0.90          & \cellcolor[HTML]{57BB8A}\textbf{0.88} & \cellcolor[HTML]{87CEAB}0.90          & \cellcolor[HTML]{FFF2CC}50.13          & \cellcolor[HTML]{E67C73}55.57           & \cellcolor[HTML]{57BB8A}\textbf{50.04} & \cellcolor[HTML]{EC9C95}54.27           & \cellcolor[HTML]{FFF2CC}39.13          & \cellcolor[HTML]{E67C73}41.01          & \cellcolor[HTML]{57BB8A}\textbf{39.04} & \cellcolor[HTML]{FBE6E4}39.50          \\ \cline{2-14} 
\multicolumn{1}{l|}{\multirow{-7}{*}{H36M}}                                                    & \multicolumn{1}{l|}{\begin{tabular}[c]{@{}l@{}}VideoPose\\ (T=81)\end{tabular}}            & \cellcolor[HTML]{FFF2CC}3.06  & \cellcolor[HTML]{78C8A1}0.88          & \cellcolor[HTML]{57BB8A}\textbf{0.87} & \cellcolor[HTML]{9AD6B8}0.89          & \cellcolor[HTML]{FFF2CC}48.97          & \cellcolor[HTML]{E67C73}54.64           & \cellcolor[HTML]{57BB8A}\textbf{48.89} & \cellcolor[HTML]{ED9C96}53.26           & \cellcolor[HTML]{FFF2CC}38.27          & \cellcolor[HTML]{E67C73}40.27          & \cellcolor[HTML]{57BB8A}\textbf{38.21} & \cellcolor[HTML]{FAE1DF}38.74          \\ \hline
\multicolumn{1}{l|}{}                                                                          & \multicolumn{1}{l|}{SPIN}                                                                  & \cellcolor[HTML]{FFF2CC}28.54 & \cellcolor[HTML]{A7DBC1}6.42          & \cellcolor[HTML]{A0D8BC}\textbf{6.41} & \cellcolor[HTML]{FFFFFF}6.54          & \cellcolor[HTML]{FFF2CC}100.74         & \cellcolor[HTML]{76C79F}94.35           & \cellcolor[HTML]{E67C73}101.76         & \cellcolor[HTML]{57BB8A}\textbf{92.89}  & \cellcolor[HTML]{FFF2CC}61.35          & \cellcolor[HTML]{E67C73}62.9           & \cellcolor[HTML]{ED9D97}62.51          & \cellcolor[HTML]{57BB8A}\textbf{60.22} \\ \cline{2-14} 
\multicolumn{1}{l|}{}                                                                          & \multicolumn{1}{l|}{TCMR}                                                                  & \cellcolor[HTML]{FFF2CC}7.92  & \cellcolor[HTML]{B7E1CC}6.47          & \cellcolor[HTML]{57BB8A}\textbf{6.45} & \cellcolor[HTML]{EBF7F1}6.49          & \cellcolor[HTML]{FFF2CC}92.83          & \cellcolor[HTML]{C3E6D5}91.45           & \cellcolor[HTML]{E67C73}93.79          & \cellcolor[HTML]{57BB8A}\textbf{88.93}  & \cellcolor[HTML]{FFF2CC}\textbf{58.16} & \cellcolor[HTML]{E67C73}60.57          & \cellcolor[HTML]{FBE7E5}59.21          & \cellcolor[HTML]{FFFFFF}58.89          \\ \cline{2-14} 
\multicolumn{1}{l|}{\multirow{-3}{*}{\begin{tabular}[c]{@{}l@{}}MPI-INF\\ -3DHP\end{tabular}}} & \multicolumn{1}{l|}{VIBE}                                                                  & \cellcolor[HTML]{FFF2CC}22.37 & \cellcolor[HTML]{57BB8A}\textbf{6.37} & \cellcolor[HTML]{B7E1CC}6.38          & \cellcolor[HTML]{FFFFFF}6.50          & \cellcolor[HTML]{FFF2CC}92.39          & \cellcolor[HTML]{5FBE8F}87.81           & \cellcolor[HTML]{E67C73}93.33          & \cellcolor[HTML]{57BB8A}\textbf{87.57}  & \cellcolor[HTML]{FFF2CC}58.85          & \cellcolor[HTML]{E67C73}60.03          & \cellcolor[HTML]{E98A82}59.91          & \cellcolor[HTML]{57BB8A}\textbf{58.84} \\ \hline
\multicolumn{1}{l|}{}                                                                          & \multicolumn{1}{l|}{TposeNet}                                                              & \cellcolor[HTML]{FFF2CC}12.70 & \cellcolor[HTML]{57BB8A}\textbf{7.23} & \cellcolor[HTML]{57BB8A}\textbf{7.23} & \cellcolor[HTML]{ABDDC4}7.29          & \cellcolor[HTML]{FFF2CC}103.33         & \cellcolor[HTML]{57BB8A}100.78          & \cellcolor[HTML]{FFFFFF}102.08         & \cellcolor[HTML]{57BB8A}\textbf{101.21} & \cellcolor[HTML]{FFF2CC}68.36          & \cellcolor[HTML]{B0DFC8}68.10          & \cellcolor[HTML]{57BB8A}\textbf{67.80} & \cellcolor[HTML]{E67C73}68.38          \\ \cline{2-14} 
\multicolumn{1}{l|}{\multirow{-2}{*}{MuPoTS}}                                                  & \multicolumn{1}{l|}{\begin{tabular}[c]{@{}l@{}}TposeNet \\ w/\\ RefineNet\end{tabular}}    & \cellcolor[HTML]{FFF2CC}9.53  & \cellcolor[HTML]{72C69D}7.21          & \cellcolor[HTML]{57BB8A}\textbf{7.20} & \cellcolor[HTML]{ABDDC4}7.23          & \cellcolor[HTML]{FFF2CC}93.97          & \cellcolor[HTML]{57BB8A}\textbf{91.78}  & \cellcolor[HTML]{FFFFFF}93.34          & \cellcolor[HTML]{71C59C}92.03           & \cellcolor[HTML]{FFF2CC}65.16          & \cellcolor[HTML]{CCEADB}65.06          & \cellcolor[HTML]{57BB8A}\textbf{64.83} & \cellcolor[HTML]{E67C73}65.33          \\ \hline
\multicolumn{1}{l|}{}                                                                          & \multicolumn{1}{l|}{EFT}                                                                   & \cellcolor[HTML]{FFF2CC}29.03 & \cellcolor[HTML]{57BB8A}\textbf{5.30} & \cellcolor[HTML]{57BB8A}\textbf{5.30} & \cellcolor[HTML]{A8DBC2}5.44          & \cellcolor[HTML]{FFF2CC}81.60          & \cellcolor[HTML]{86CEAB}79.79           & \cellcolor[HTML]{FFFFFF}80.57          & \cellcolor[HTML]{57BB8A}\textbf{79.48}  & \cellcolor[HTML]{FFF2CC}48.78          & \cellcolor[HTML]{E67C73}49.08          & \cellcolor[HTML]{F5C7C3}48.91          & \cellcolor[HTML]{57BB8A}\textbf{48.54} \\ \cline{2-14} 
\multicolumn{1}{l|}{}                                                                          & \multicolumn{1}{l|}{PARE}                                                                  & \cellcolor[HTML]{FFF2CC}22.77 & \cellcolor[HTML]{69C297}5.24          & \cellcolor[HTML]{57BB8A}\textbf{5.22} & \cellcolor[HTML]{AADCC4}5.31          & \cellcolor[HTML]{FFF2CC}71.80          & \cellcolor[HTML]{E67C73}72.27           & \cellcolor[HTML]{93D3B4}71.36          & \cellcolor[HTML]{57BB8A}\textbf{71.11}  & \cellcolor[HTML]{FFF2CC}\textbf{43.44} & \cellcolor[HTML]{E67C73}45.18          & \cellcolor[HTML]{FFFFFF}43.49          & \cellcolor[HTML]{FEF9F8}43.58          \\ \cline{2-14} 
\multicolumn{1}{l|}{}                                                                          & \multicolumn{1}{l|}{SPIN}                                                                  & \cellcolor[HTML]{FFF2CC}30.84 & \cellcolor[HTML]{5BBC8D}5.36          & \cellcolor[HTML]{57BB8A}\textbf{5.35} & \cellcolor[HTML]{ADDDC6}5.53          & \cellcolor[HTML]{FFF2CC}87.58          & \cellcolor[HTML]{E67C73}87.99           & \cellcolor[HTML]{57BB8A}\textbf{85.60} & \cellcolor[HTML]{B1DFC9}86.67           & \cellcolor[HTML]{FFF2CC}53.28          & \cellcolor[HTML]{E67C73}54.42          & \cellcolor[HTML]{C0E5D3}53.08          & \cellcolor[HTML]{57BB8A}\textbf{52.74} \\ \cline{2-14} 
\multicolumn{1}{l|}{}                                                                          & \multicolumn{1}{l|}{TCMR}                                                                  & \cellcolor[HTML]{FFF2CC}6.76  & \cellcolor[HTML]{8ED1B0}5.96          & \cellcolor[HTML]{57BB8A}\textbf{5.95} & \cellcolor[HTML]{8ED1B0}5.96          & \cellcolor[HTML]{FFF2CC}\textbf{86.46} & \cellcolor[HTML]{E67C73}87.70           & \cellcolor[HTML]{FFFFFF}86.48          & \cellcolor[HTML]{E8857D}87.62           & \cellcolor[HTML]{FFF2CC}\textbf{52.67} & \cellcolor[HTML]{E67C73}53.61          & \cellcolor[HTML]{FFFFFF}53.00          & \cellcolor[HTML]{EFA7A1}53.41          \\ \cline{2-11} \cline{13-14} 
\multicolumn{1}{l|}{\multirow{-5}{*}{3DPW}}                                                    & \multicolumn{1}{l|}{VIBE}                                                                  & \cellcolor[HTML]{FFF2CC}23.16 & \cellcolor[HTML]{57BB8A}\textbf{5.98} & \cellcolor[HTML]{57BB8A}\textbf{5.98} & \cellcolor[HTML]{B1DFC8}6.12          & \cellcolor[HTML]{FFF2CC}82.97          & \cellcolor[HTML]{E67C73}85.89           & \cellcolor[HTML]{57BB8A}\textbf{81.49} & \cellcolor[HTML]{F3C0BB}84.39           & \cellcolor[HTML]{FFF2CC}52.00          & \cellcolor[HTML]{E67C73}52.49          & \cellcolor[HTML]{57BB8A}\textbf{51.70} & \cellcolor[HTML]{FCEFEE}52.06          \\ \hline
\end{tabular}
\label{tab:supp_gd_3d}
\end{table}

\renewcommand\arraystretch{1}

\subsection{Smoothness on Synthetic Data}
Due to the lack of pairwise labeled data, some approaches ~\cite{gauss2021spsmoothing} for Mocap sensors denoising verify the validity of their approaches on synthetic noise, like Gaussian noises.
We follow their methods to generate the noisy poses, adding different levels of Gaussian noises on the ground truth data. We take the Human3.6M dataset as an example. In the training stage, we generate Gaussian noises with the probability $p$ and noise variance $\sigma$ on the ground truth 2D or 3D positions for 2D or 3D pose estimation respectively as synthetic training data. \modelname~can be trained on these synthetic data. In the inference stage, we also add the same noise level to the testing set as the synthetic test data. Table~\ref{tab:hm_gau} gives the corresponding results of our model. 
\modelname~can refine the noises/jitters at a large margin without any spatial correlations since it utilizes the smoothness prior of human motions. For instance, in terms of 3D pose estimation, either as the variance of Gaussian noises increase from 10mm to 100mm or the probability changes from $0.1$ to $0.9$, \modelname~can decrease \textit{Accel} and \textit{MPJPE} at a large margin. Those results indicate \modelname~will be also beneficial to remove different synthetic noises. 

\begin{table}[h]
	\centering
    \small
    \caption{Comparison of the 3D pose with different synthetic noises from \textit{Gaussian Noise} on Human3.6M. $p$ is the probability of adding noise, and $\sigma$ means the variance. pix. is the abbreviation of the pixel.}
	\scriptsize
	{%
		\begin{tabular}{ll|cc|cc}
		    
			\specialrule{.1em}{.05em}{.05em}

			&Gaussion Noise & In \textit{Accel}& Out \textit{Accel}& In \textit{MPJPE}&Out \textit{MPJPE}\\
            \midrule
 
			\parbox{2.5mm}{\multirow{6}{*}{\rotatebox[origin=c]{90}{2D Pose}}}
			& $p$ = 0.5, $\sigma$ = 10 pix. &10.10&\textbf{0.20}&3.56&\textbf{0.83}\\
            & $p$ = 0.5, $\sigma$ = 50 pix.&50.53&\textbf{0.35}&17.80 &\textbf{2.02}\\
            &$p$ = 0.5, $\sigma$ = 100 pix. &101.06&\textbf{0.31}&35.59&\textbf{1.42}\\
            \cmidrule{2-6}
            
            &$p$ = 0.1, $\sigma$ = 50 pix.&14.31&\textbf{0.19}&3.90&\textbf{0.67}\\
            & $p$ = 0.5, $\sigma$ = 50 pix.&50.53&\textbf{0.35}&17.80 &\textbf{2.02}\\
            & $p$ = 0.9, $\sigma$ = 50 pix.&72.26&\textbf{0.57}&28.97 &\textbf{6.00}\\
            
		    \midrule
			
		    \parbox{2.5mm}{\multirow{6}{*}{\rotatebox[origin=c]{90}{3D Pose}}} 
			& $p$ = 0.5, $\sigma$ = 10mm &26.25&\textbf{0.84}&9.68&\textbf{3.54}\\
            & $p$ = 0.5, $\sigma$ = 50mm&131.25&\textbf{1.55}& 48.42&\textbf{7.00}\\
            &$p$ = 0.5, $\sigma$ = 100mm &262.49&\textbf{1.24}&96.84&\textbf{20.38}\\
            \cmidrule{2-6}
            &$p$ = 0.1, $\sigma$ = 50mm&40.68&\textbf{1.03}&11.46&\textbf{2.46}\\
            & $p$ = 0.5, $\sigma$ = 50mm&131.25&\textbf{1.55}& 48.42&\textbf{7.00}\\
            & $p$ = 0.9, $\sigma$ = 50mm & 184.32&\textbf{2.10}&74.46 &\textbf{16.85}\\
        \midrule
        \end{tabular}%
	}
	\label{tab:hm_gau}
\end{table}

\subsection{More Ablation Study}

\noindent\textbf{Impact on Loss Function.}
As mentioned in the main paper Sec 4.3, we use $L_{pose}$+$L_{acc}$ as our final objective function. Here we explore how the loss functions affect the performance in Table~\ref{tab:loss}. First, we find that only single-frame supervision $L_{pose}$ would be slightly worse than our result by $5.51$\% in \textit{Accel}, while the \emph{MPJPE}s are competitive. It shows the precision can be optimized well by the $L_{pose}$. Next, only with $L_{acc}$ will make all results worst, indicating the significant necessity of $L_{pose}$ supervision. Last, adding $L_{pose}$ and $L_{acc}$ together to train the \modelname~will benefit both smoothness and precision, proving that $L_{acc}$ companies with $L_{pose}$ can play its smooth role.

\begin{table}[h]
	\centering
    \small
    \caption{Comparison of refined results by different loss functions based on the outputs of the SMPL-based method EFT~\cite{joo2020eft} on the 3DPW dataset. }
	\scriptsize
	{%
		\begin{tabular}{l|ccc}
			\specialrule{.1em}{.05em}{.05em}
			Method& \textit{Accel}& \emph{MPJPE}& \emph{PA-MPJPE}\\
			\midrule
            EFT &32.71&90.32&52.19\\
            \midrule
            $L_{pose}$  &6.42&86.63&\textbf{50.82}\\
            $L_{acc}$ &7.63&446.54&356.61\\
            $L_{pose}$+$L_{acc}$&\textbf{6.30}&\textbf{86.39}&50.60\\
        \midrule
        \end{tabular}%
	}
	\label{tab:loss}
\end{table}

\noindent\textbf{Impact on Motion Modalities.}
Motivated by this natural smoothness characteristic, we can unify various continuous modalities and make \modelname~generalize well across them. 
In particular, 2D, 3D positions, and 6D rotation matrices are continuous modalities of the same space in neural networks. In contrast, the rotation representations as axis-angle or quaternion are discontinuous in the real Euclidean spaces~\cite{Zhou2019OnTC}, which may be hard for neural networks to learn. 
Accordingly, we explore the effects of these modalities used to train \modelname~on EFT~\cite{joo2020eft}. 
Table~\ref{tab:3dpw_represent} shows the training results on each motion modality. We can see that the axis-angle or quaternion obtains worse results on both smoothness and precision. They may encounter some sudden changes/flips leading to poor results due to the discontinuity of the expression.
Instead, the 6D rotation matrix and 3D position will be more suitable to learn and improve all metrics.
Furthermore, 3D positions reach the best performance by decreasing $82.15$\% in \textit{Accel}, $5.72$\% in \textit{MPJPE}, and $3.60$\% in \textit{PA-MPJPE}. 

\begin{table}[h]
	\centering
    \small
    \caption{Comparison of refined results trained by different motion modalities based on the outputs of EFT~\cite{joo2020eft} on the 3DPW dataset. }
	\scriptsize
	{%
		\begin{tabular}{l|ccc}
			\specialrule{.1em}{.05em}{.05em}
			Method& \textit{Accel}& \emph{MPJPE}& \emph{PA-MPJPE}\\
			\midrule
            EFT~\cite{joo2020eft} &32.71&90.32&52.19\\
            \midrule
            Angle-Axis&77.89 &172.17&51.38\\
            Quaternion &28.50&91.23&51.03\\
            6D Rotation &6.43&86.92&50.87\\
            3D Position&\textbf{6.30}&\textbf{86.39}&\textbf{50.60}\\
        \midrule
        \end{tabular}%
	}
	\label{tab:3dpw_represent}
\end{table}

\begin{figure*}[h]
\begin{center}
\includegraphics[width=0.96\textwidth]{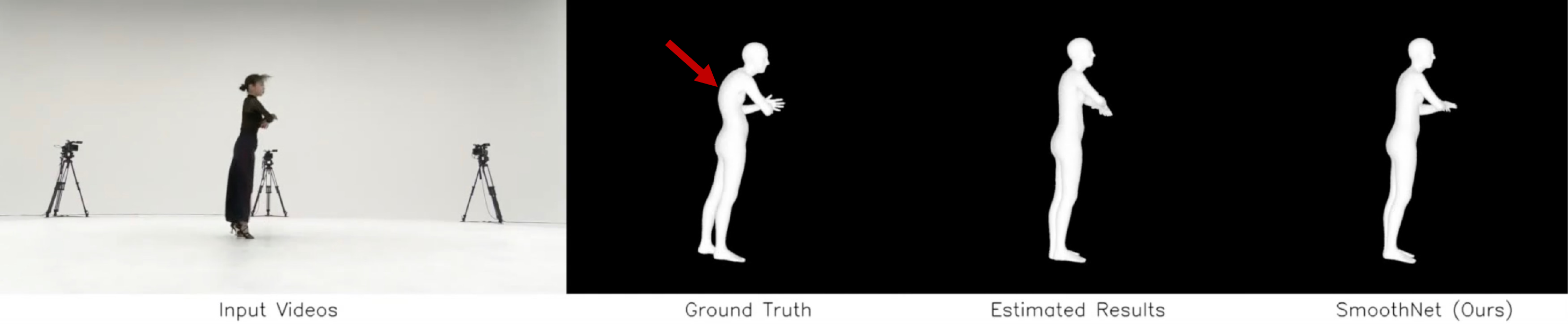}
\end{center}
\caption{Comparison the results of the ground truth, VIBE~\cite{kocabas2020vibe} with VIBE w/ \modelname~on AIST++ dataset.}
\label{fig:aist}
\end{figure*}

Last, to explore whether there is also better generalization between different continuous modalities, such as 3D position and 6D rotation matrix, cross-modality tests were carried out demonstrated in Table~\ref{tab:cross_represent}. We can summarize these observations: (i) when tested across modalities, all results will be worse relative to the modality the model trained on; (ii) \modelname~trained in 3D positions, smoothed directly over the representation of the 6D rotation matrix, can achieve even better performance than training on the 6D rotation matrix itself. Hence, these results motivate us to use 3D positions as supervision by default, where 3D positions contain more information than 2D positions, and their ground-truth are usually more precise than the 6D rotation matrix (explicitly found in the AIST++ dataset, like Figure~\ref{fig:aist}).

\begin{table}[h]
	\centering
    \small
    \caption{Comparison of refined results by \textit{cross motion representations testing} based on the outputs of EFT~\cite{joo2020eft} on the 3DPW dataset. \textit{Cross-Test} means training the \modelname~on a motion representation while testing it on another modality directly.}
	\scriptsize
	{%
		\begin{tabular}{l|ccc}
			\specialrule{.1em}{.05em}{.05em}
			Method& \textit{Accel}& \emph{MPJPE}& \emph{PA-MPJPE}\\
			\midrule
            EFT &32.71&90.32&52.19\\
            \midrule
            6D Rotation &\underline{6.43}&86.92&50.87\\
            Cross-Test on 3D Position&7.10&88.13&51.79\\
            \midrule
            3D Position&\textbf{6.30}&\textbf{86.39}&\textbf{50.60}\\
            Cross-Test on 6D Rotation &6.47&\underline{86.82}&\underline{50.81}\\
        \midrule
        \end{tabular}%
	}
	\label{tab:cross_represent}
\end{table}

\noindent\textbf{Effect of Normalization Strategies.}
Normalization is an effective way to calibrate biased errors and improve the generalization ability. As a plug-and-play network, we also explore how different normalization strategies influence the results, especially the generalization ability. In the main paper, we do not use any normalization by default.

We adopt three normalization strategies.
Particularly, \textit{w/o Norm.} denotes taking the original estimated results without normalization.
\textit{Sequence Norm.} indicates normalizing each input axis $\mathbf{\hat{Y}}_{i}$ with means and variances computed from input sequences along the axis.
Because the estimated inputs are always noisy and the bias shift between the training data and testing data, the above normalization methods will be affected. Instead, using the mean and variance from the ground truth (with $\dagger$) along each axis can avoid such influences and we can explore the upper bound performance under the \textit{Sequence Norm.} normalization.

\begin{table}[h]
    \vspace{-0.5cm}
	\centering
    \small
    \caption{Comparison of the results of different normalization based on the outputs of EFT~\cite{joo2020eft} and \textit{cross-backbone testing} on the outputs of TCMR~\cite{choi2021tcmr} on 3DPW dataset. $\dagger$ means using the same mean and variance as the ground truth to explore the upper bound performance.}
	\scriptsize
	{%
		\begin{tabular}{l|ccc}
		    
			\specialrule{.1em}{.05em}{.05em}
			Method& \textit{Accel}& \emph{MPJPE}& \emph{PA-MPJPE}\\
			
			\midrule
            
            EFT~\cite{joo2020eft} &32.71&90.32&52.19\\
            \midrule
            w/o Norm. &\textbf{5.80}&85.16&50.31\\
            Sequence Norm. &5.82&88.21&51.06\\
            Sequence Norm. $\dagger$ &\textbf{5.80}&\textbf{61.65}&\textbf{44.28}\\
            \midrule
            TCMR~\cite{choi2021tcmr} &6.76 &86.46 &52.67 \\
            \midrule
            w/o Norm. &\textbf{5.91}& 86.04 &52.42\\
            Sequence Norm &6.00&86.34&52.87\\
            Sequence Norm $\dagger$&5.92&\textbf{68.51}&\textbf{49.15}\\
            \midrule
        \end{tabular}%
	}
	\label{tab:upper_bound}
\vspace{-0.5cm}
\end{table}

In Table~\ref{tab:upper_bound}, we compare the performance of different normalizations based on the outputs of EFT~\cite{joo2020eft} on the 3DPW dataset in the upper table. We can discover that the smoothing ability for all normalizations is similar, and the main difference lies in the degree of biased error removal. To be specific, under the \textit{Sequence Norm. $\dagger$} normalization, the \emph{MPJPE} can decrease from $85.16$mm to $61.65$mm, improved by $27.5$\%. To explore the generalization ability across backbones, we further test \modelname~trained on EFT-3DPW on TCMR~\cite{choi2021tcmr}-3DPW. From the lower part of the table, we can get similar conclusions as above. In specific, \modelname~can reduce \textit{Accel} from $6.77mm/frame^2$ to about $6mm/frame^2$, and the upper bound of \emph{MPJPE} can be $68.51mm$ (improvement by $20.8$\%) from the refinement stage.

\section{Qualitative Results}
\label{sec:supp_viz}
As jitters seriously affect visual effect, we visualize the results from several tasks, such as 2D pose estimation, 3D pose estimation, and model-based body recovery. For 2D and 3D pose estimation, we show two kinds of actions on Human3.6M respectively with the corresponding \textit{Accel} and \textit{MPJPE} for each frame. The estimated 2D poses are from the single-frame SOTA method RLE~\cite{li2021rle}, and the estimated 3D poses are from the single-frame method FCN~\cite{martinez2017simple}. 
For model-based methods, the estimated results come from VIBE~\cite{kocabas2020vibe}
on AIST++ dataset and SPIN~\cite{kolotouros2019spin} on 3DPW dataset.

We can observe that the jitters in a video are highly-unbalanced, where most frames suffer from slight jitters while long-term significant jitters will be accompanied by large biased errors. 
\modelname~can relieve not only small jitters but long-term jitters well. And it can boost both smoothness and precision significantly. 
Specifically, unlike low-pass filters~\cite{gauss2021spsmoothing,press1990savitzky,hyndman2011moving}, our method can estimate the high-frequency movements well, like the action \textit{Posing}.
Finally, we observe that the ground-truth 6D rotation matrices from AIST++ is not quite accurate, as the SMPL annotations are fitted with few constraints. For example, the red arrow in Figure~\ref{fig:aist} illustrates that a SMPL fitting from AIST++ has a bulging back problem. Instead, their $14$ skeletal 3D positions are more precise. When it comes to model training, the quality of annotation is crucial for the success of a data-driven model.
As \modelname~is devised to operate on temporal axis, it is capable of training on one modality and testing on the other, so as to have the flexibility of choosing more precise annotated modality for training. This property makes \modelname~applicable to datasets with different annotation qualities from different modalities.


\bibliographystyle{splncs04}